\documentclass{article}

 \usepackage[preprint, nonatbib]{neurips_2019}

\usepackage[utf8]{inputenc} 
\usepackage[T1]{fontenc}    
\usepackage{hyperref}       
\usepackage{url}            
\usepackage{booktabs}       
\usepackage{amsfonts}       
\usepackage{amsthm}         
\usepackage{nicefrac}       
\usepackage{microtype}      
\usepackage{color}
\usepackage{algorithm}      
\usepackage[noend]{algpseudocode} 
\usepackage[pdftex]{graphicx}
\usepackage[export]{adjustbox}
\usepackage{subfig}
\usepackage{xr}
\usepackage{amsmath}
\usepackage{amssymb}
\newcommand{\wail}{\textsc{Wail}}
\newcommand{\gail}{\textsc{Gail}}
\newcommand{\bc}{\textsc{Bc}}
\newcommand{\al}{\textsc{Al}}
\newcommand{\irl}{\textsc{Irl}}
\newcommand{\rl}{\textsc{Rl}}
\usepackage{mathtools}
\newcommand{\defeq}{\vcentcolon=}

\def\cond{\,|\,}
\newtheorem{theorem}{Theorem}[section]
\newtheorem{proposition}{Proposition}[section]

\newcommand{\ie}{\emph{i.e.~}}
\newcommand{\eg}{\emph{e.g.~}}

\newcommand{\set}[1]{\ensuremath{\mathcal{#1}}}

\newcommand{\bb}[1]{\ensuremath{\mathbb{#1}}}

\newcommand{\argmax}{\operatornamewithlimits{\arg\,\max}}

\newcommand{\argmin}{\operatornamewithlimits{\arg\,\min}}

\DeclareMathOperator{\RL}{RL}
\DeclareMathOperator{\IRL}{IRL}

\renewcommand{\eqref}[1]{Eq.(\ref{#1})}
\renewcommand{\algref}[1]{Algorithm (\ref{#1})}
\newcommand{\figref}[1]{Fig. (\ref{#1})}
\newcommand{\tblref}[1]{Table. (\ref{#1})}

\newcommand{\SA}{\mathcal{S}\times\mathcal{A}}
\newcommand{\cS}{\mathcal{S}}
\newcommand{\cA}{\mathcal{A}}
\newcommand{\cR}{\mathcal{R}}
\newcommand{\cB}{\mathcal{B}}
\renewcommand{\P}{\mathbb{P}}
\newcommand{\E}{\mathbb{E}}
\newcommand{\R}{\mathbb{R}}

\begin{document}

\title{Wasserstein Adversarial Imitation Learning}

%

\author{%
	Huang Xiao \\
	\texttt{Huang.Xiao@de.bosch.com}
	\And 
    Michael Herman \\
	\texttt{Michael.Herman@de.bosch.com}
	\And 
	Joerg Wagner \\
	\texttt{Joerg.Wagner3@de.bosch.com}
	\And
	Sebastian Ziesche \\
	\texttt{Sebastian.Ziesche@de.bosch.com}
	\And
	Jalal Etesami \\
	\texttt{Jalal.Etsami@de.bosch.com}
	\And
	Thai Hong Linh \\
	\texttt{HongLinh.Thai@de.bosch.com}
	\And
		\text{ \textnormal{Bosch Center for Artificial Intelligence}}\\
		Robert Bosch GmbH\\
		Rennigen, Stuttgart in Germany \\
}

\maketitle
\begin{abstract}
	Imitation Learning describes the problem of recovering an expert policy from demonstrations. While inverse reinforcement learning approaches are known to be very sample-efficient in terms of expert demonstrations, they usually require problem-dependent reward functions or a (task-)specific reward-function regularization. In this paper, we show a natural connection between inverse reinforcement learning approaches and Optimal Transport, that enables more general reward functions with desirable properties (\eg smoothness). Based on our observation, we propose a novel approach called Wasserstein Adversarial Imitation Learning. Our approach considers the Kantorovich potentials as a reward function and further leverages regularized optimal transport to enable large-scale applications. In several robotic experiments, our approach outperforms the baselines in terms of average cumulative rewards and shows a significant improvement in sample-efficiency, by requiring just one expert demonstration.
\end{abstract}

\section{Introduction}
The increasing capabilities of autonomous systems allow to apply them in more and more complex environments. However, a fast and easy deployment requires simple programming approaches to adjust autonomous systems to new goals and tasks. A well-known class of approaches offering such properties is apprenticeship learning (\al, learning from demonstration) \cite{Argall2009SurveyLfD}, which summarizes the methods for teaching new skills by demonstration, instead of programming them directly. Common \al\ methods include, for instances, behavioral cloning (\textsc{Bc}) \cite{Bain1999BehavioralCloning}, \al\ via inverse reinforcement learning (\textsc{Irl}) \cite{Russell1998LearningAgents}, or adversarial imitation learning \cite{GAIL}.

The goal of behavior cloning is to mimic the expert's behavior by estimating the policy directly from demonstrations. This method typically requires a large amount of data, and the resulting models often suffer from compounding errors due to covariate shifts caused by policy errors \cite{ERILross10a}. Instead of copying the behavior directly, inverse reinforcement learning assumes rational agents to estimate an unknown reward function that represents the underlying motivations and goals. Hence, the reward function is considered to be a more parsimonious \cite{Ng2000AlgorithmsForIRL} or succinct \cite{Abbeel:2004:ALV:1015330.1015430} description of the expert's objective. It allows to infer optimal actions for unobserved states, new environments, or various dynamics. Although \irl\ has shown to be very sample-efficient in terms of expert demonstrations (\cite{Abbeel2008ParkingLotNavigation,Kuderer2013TeachingMobileRobots}), it often requires careful hand-engineering of reward functions, as the \irl\ problem is known to be ill-posed and multiple reward functions can explain a certain observed behavior. Furthermore, many \irl\ approaches require solving the reinforcement learning (\rl) problem in an inner loop, which becomes increasingly prohibitive in high-dimensional continuous space.


To overcome these problems, generative adversarial imitation learning (\gail) \cite{GAIL} proposes a \textsc{Gan}-based method to learn the policy directly by matching occupancy measures, which is shown to be similar as running \rl\ after \irl, and therefore preserves the advantages of \irl\ while being computationally more viable. However, the proposed method intrinsically minimizes Jensen-Shannon divergence by learning a discriminator, which cannot be interpreted as a proper reward function eventually. Recently, adversarial IRL \cite{fu2018learning} decouples the discriminator to output a valid reward function, however, a specific network architecture is required. Nevertheless, it is known that the standard \textsc{Gan} training \cite{Goodfellow:2014:GAN} is prone to training instabilities, \eg mode collapse or vanishing gradient. Hence researchers investigate several alternative approaches to replace the Jensen-Shannon divergence objective, most notably, by a Wasserstein \textsc{Gan}-based objective (\cite{Wang_WIOC,tail2018socially}). While the quantative results indicate an improvement in performance, a theoretical justification is missing so far.

In this work, we show that it is indeed possible to generalise the existing approaches to a larger set of reward function space. In particular, we derive a novel imitation learning algorithm built on the Wasserstein distance. Our contribution is threefold: first, we justify theoretically that there is a natural connection between apprenticeship learning by minimizing  
Integral Probability Metrics, \eg Wasserstein distance \cite{sriperumbudur2012empirical},  
and via inverse reinforcement learning. This enables a broader class of reward functions with desirable properties, \eg smoothness. When choosing Wasserstein distance, we observe strong connection to the optimal transport (OT) theory that the Kantorovich potential in the dual form of OT problem can be interpreted as a valid reward function.
Second, we propose a novel approach called Wasserstein Adversarial Imitation Learning (\wail), which leverages regularized optimal transport to enable large-scale applications. Finally, we perform several robotic experiments, in which our approach outperforms the baselines in terms of average cumulative rewards and shows a significant improvement in sample-efficiency, by requiring just one expert demonstration.

\section{Background}
\label{sec:background}
\textbf{Preliminaries. }
An infinite horizon discounted Markov decision process setting $\set{M}$ is defined by the tuple $(\set{S}, \set{A}, p, r, \mu_0, \gamma)$ consisting of the finite state space $\set{S}$, the finite action space $\set{A}$ and the transition function $p(s'\mid s, a)$ as being a probability measure on $\cS$ for all $(s, a) \in \SA$. Moreover, we have the reward function $r:\SA \to \R$ (\ie, a function mapping from $\SA$ to $\R$), the starting distribution $\mu_0$ on $\cS$ satisfying $\mu_0(s) > 0$ for all $s \in \cS$ and finally the discount factor $\gamma \in (0, 1)$. 

If we combine this setting with a stochastic policy $\pi$ from the set of policies $\Pi$, i.e. a conditional probability distribution on $\cA$ given some state $s \in \cS$, we obtain a Markov chain $M_\pi = (S_0, A_0, S_1, A_1, \dots)$ in the following natural way: Take a random starting state $s \sim \mu_0$, choose a random action $a \sim \pi(\cdot \mid s)$ and then restart the chain with probability $1-\gamma$ or choose the next state $s' \sim p(\cdot \mid s, a)$ otherwise, repeat the last two steps. 

It is well known, that $M_\pi$, if seen as a Markov chain on $\SA$, has a stationary distribution (or occupancy measure) $\rho_\pi$ which satisfies $\rho_\pi(s, a) = (1-\gamma) \pi(a \mid s) \sum_{t = 0}^{\infty} \gamma^t \P[S_t = s]$ as well as the Bellmann equation $\sum_{a' \in \cA} \rho(s', a') = (1 - \gamma)\mu_0(s') + \gamma \sum_{(s, a) \in \SA} p(s' \mid s, a) \rho(s, a)$ for all $s, s' \in \cS$ and $a \in \cA$. Moreover, there is a one-to-one correspondence between those measures and the policies in $\Pi$.

\begin{proposition}{(see Theorem 2. of \cite{Syed:2008:ALU:1390156.1390286}))}
	The mapping $\rho \mapsto \pi_\rho$ defined by $\pi_\rho(a \mid s) := \rho(s, a) / \sum_{a' \in \cA} \rho(s, a')$ is a bijection between $\Pi$ and the set $\cB$ of measures on $\SA$ satisfying the Bellmann equation.
\end{proposition}

Due to this correspondence, we write $\E_{\pi}[X(s, a)] := \E_{\rho_\pi}[X(s, a)] := \E_{(s, a) \sim \rho_\pi}[X(s, a)]$ for the expected value of a random variable $X$ on $\SA$ with respect to $\rho_\pi$. We observe, that the expected cumulative reward of $M_\pi$, \ie the expected sum of rewards $r(s_t, a_t)$ up to the first restart of the chain, is given by $\E[\sum_{t = 0}^{\infty} \gamma^t r(s_t, a_t)]$. Hence it is easy to derive that it is also equal to $\E_{\pi}[r(s, a)] / (1 - \gamma)$. For brevity, we write $\langle r, \rho \rangle := \sum_{(s, a) \in \SA} r(s, a) \rho(s, a) $ sometimes.

\textbf{Inverse Reinforcement Learning. } The goal of reinforcement learning is to learn a policy $\pi$ that maximizes the expected cumulative rewards. Hence, typical \rl\ approaches assume the reward function $r$ to be given. However, for many problems the true reward function is not known or too hard to specify. Therefore, \irl\ resorts to estiamte the unknown reward function from the expert policy $\pi_E$ or demonstrations. Morever, to weaken the assumption about the optimality of the policy, maximum causal entropy \irl\ (\textsc{Mce-Irl}) \cite{Ziebart2010Modeling} has been proposed, which learns a reward function $r \in \cR$ as 
\begin{align}
	\max_{r \in \cR} \left(\min_{\pi \in \Pi} - H(\pi) - \bb E_{\pi}\left[r(s,a)\right] \right) + \bb E_{\pi_E}\left[r(s,a)\right] 
	\label{eq:MCEIRL}
\end{align}
where $H(\pi) := \E_{(s, a) \sim \rho_\pi}[-\log \pi(a \mid s)] / (1 - \gamma)$ is the $\gamma$-discounted causal entropy of the policy \cite{ZhouMCEIRL}. In practice, the full expert policy is often not available, while it is possible to query expert demonstrations. Hence, the expected expert rewards are often approximated via a finite set of demonstrations. Then, \textsc{Mce-Irl} seeks for a reward function that assigns high rewards to expert demonstrations and low rewards to all the others, in  
favor of high entropy policies. The corresponding \rl\ problem follows to derive a policy from this reward function: $\min_{\pi \in \Pi} - H(\pi) - \bb E_{\pi}\left[r(s,a)\right]$, which is embedded in the inner loop of \irl.

\textbf{Generative Adversarial Imitation Learning. } \textsc{Irl}-based imitation learning approaches have shown to generalise well, if the estimated reward functions are properly designed. However, many approaches are inefficient, as they require to solve the \rl\ problem in an inner loop. In addition, the goal of imitation learning is typically an imitation policy. Hence, the reward function from the intermediate \irl\ step is often not required. Previous work \cite{GAIL} has characterized the policies that originate from running \rl\ after \irl, and propose to extend the \irl\ problem from \eqref{eq:MCEIRL} by imposing an additional cost function regularizer $\psi(c)$,
\begin{align}
\IRL_\psi(\pi_E) = \max_{c \in \R^{\SA}} - \psi(c) + \left(\min_{\pi \in \Pi} - H(\pi) + \bb E_{\pi}\left[c(s,a)\right] \right) - \bb E_{\pi_E}\left[c(s,a)\right],
\label{eq:GAIL_IRL}
\end{align}
with cost functions $c \in \R^{\SA}$.
Based on this definition, the combined optimization problem of learning a policy via \rl\ with the learned reward function from \irl\ results in:
\begin{align}
\RL \circ \IRL_\psi(\pi_E) = \argmin_{\pi \in \Pi} - H(\pi) + \psi^\ast \left(\rho_\pi - \rho_{\pi_E} \right)
\label{eq:GAIL_Prop32}
\end{align}
where $\psi^\ast$ is the convex conjugate of the cost function regularizer $\psi(c)$ that measures a distance between the induced occupancy measures of the expert and the learned policy. In practice, a discriminator $D(s,a)$ is employed to differentiate the state-actions from both the expert and learned policy. And the distance measure is formulated as: $\bb E_{(s,a)\sim \rho_{\pi_E}}\left\lbrack \log(1-D(s,a))\right\rbrack + \bb E_{(s,a)\sim\rho_\pi}\left\lbrack \log(D(s,a)) \right\rbrack$, by taking $\log(D(s,a))$ as the surrogate cost function to guide the reinforcement learning. At convergence, the discriminator can not distinguish the expert and learned policy, and classifies as $0.5$ everywhere. Therefore, it can not be used as a valid cost or reward function. Note that the negative cost can be treated as a reward function, we will use reward function representation throughout this work for consistency. 

\section{From Apprenticeship Learning to Optimal Transport}
Suppose we aim to learn an imitation policy $\pi$ that tries to recover the expert's policy $\pi_E$ with corresponding occupancy measure $\rho_E := \rho_{\pi_E}$, while achieving similar expected rewards under the unknown expert's reward function. For these purposes, apprenticeship learning approaches via \irl\ have been proposed, which learn a reward function to derive a policy. 
Similarly, we use the causal entropy regularized apprenticeship learning \cite{Ho:2016:MIL:3045390.3045681} formulation as follows,
\begin{align}
\argmin_{\pi \in \Pi} -H(\pi) + \sup_{r \in \cR} \bb E_{\pi_E}\left[r(s,a)\right] - \bb E_{\pi}\left[r(s,a)\right]
\label{eq:reg_AL}
\end{align}
As mentioned by \cite{GAIL}, most apprenticeship learning approaches often strongly restrict the reward function space $\cR$ to derive efficient methods, to enhance generalisability, and also to provide feasible solutions for the ill-posed \irl\ problem. Many approaches assume the reward function to be a linear combination of hand-engineered basis functions \cite{Abbeel:2004:ALV:1015330.1015430, Ziebart2008MaxEntIRL, Syed2007GameTheoreticAL} or learn to construct features from a large collection of component features \cite{levine2010feature}. More recent approaches use Gaussian processes \cite{Levine2011GPIRL} or deep learning-based models \cite{Finn2016GuidedCostLearning} to learn non-linear reward functions from the feature- or state-action space. However, they require careful regularization to ensure that the reward functions do not degenerate. 

\textbf{From apprenticeship learning to Wasserstein distance. } Although it is often easier to specify general properties of the desired reward functions $r \in \cR$, in practice it might not be possible to specify arbitrary reward function spaces $\cR$ and to derive efficient solutions for the corresponding apprenticeship learning problem. However, suppose the function space $\set R$ is closed under negation, \ie $r\in \set R \Rightarrow -r\in \set R$, this is true if we consider for each reward function $r\in \set R$, there exists a cost function $c\defeq -r\in \set R$. Then 
we observe that the latter part of the regularized apprenticeship learning problem in \eqref{eq:reg_AL} can be interpreted as an Integral Probability Metric (\textsc{Ipm}) \cite{sriperumbudur2012empirical},
 \[ 
 \phi_\cR(\rho_E, \rho_\pi)=
\sup_{r \in \cR} \left| \langle r, \rho_E \rangle - \langle r, \rho_\pi \rangle \right|  = 
 \sup_{r \in \cR} \langle r, \rho_E \rangle - \langle r, \rho_\pi \rangle,
 \] 
 between the induced occupancy measures $\rho_E$ and $\rho_\pi$. Depending on how $\cR$ is chosen, $\phi_\cR$ turns into different metrics for probability measures like total variation: $\cR = \left\{r: \lVert r \rVert_\infty \leq 1 \right\}$, Maximum Mean Discrepancy: $\cR = \left\{r: \lVert r \rVert_\mathcal{H} \leq 1 \right\}$ with $\|\cdot\|_{\set H}$ being the norm of a reproducing kernel Hilbert space $\set H$, and also Wasserstein distance: $\cR = \text{Lip}(1)$ being the Lipschitz-functions with constant $1$ with respect to some distance function $d$ on $\SA$.

In \rl, many approaches suffer from sparse and delayed rewards. Hence, for a subsequent \rl\ task it is beneficial to have smooth reward functions that are easier to optimize and interpret. To continue our discussion, we investigate chosing $\cR$ as a class of smooth reward functions namely those, which are Lipschitz(1)-continuous with respect to some metric $d$ on $\SA$. 

\begin{proposition}
	Given two occupancy measures $\rho_\pi$ and $\rho_E$ induced by policy and expert respectively, the causal entropy regularized apprenticeship learning (\ref{eq:reg_AL}) can be formulated as following, 
\begin{align}
  \argmin_{\pi \in \Pi} -H(\pi) + W_1^d(\rho_\pi, \rho_E), 
  \label{eq:claim}
\end{align}
where $W_1^d$ is the well-known 1-Wasserstein distance of the two measures with respect to the ground cost function $d$, which is a valid distance metric defined in the state-action space.
\label{prop:claim}
\end{proposition}

\textit{Proof.} The proof can be easily shown by treating the reward function $r(s, a)$ as the Kantorovich potentials in the dual form of optimal transport (OT) problem \cite{computationalOT}. Formally, let $x\in \set X, y\in \set Y $ denote the concatenated state-action vector $\left\lbrack s, a\right\rbrack^T$ to avoid cluterred notations and assume $\set X= \set Y$, for a certain policy $\pi$ we write the dual form of OT as 
\begin{align}
	& \sup_{(f, g)\sim \set R\times \set R} \bb E_{y \sim \rho_{E}}\left[f(y)\right] + \bb E_{x\sim\rho_{\pi}}\left[g(x)\right], \label{eq:ot_IRL} \\
	\text{s.t. }& \,\,\, f(y) + g(x) \leq d(x, y), \,\,\, \forall x\times y \sim \rho_{\pi}\times \rho_E \nonumber
\end{align}
where $f, g \in \set R$ are known as Kantorovich potentials. Moreover, if $d$ is a distance metric defined in $\set X\times\set Y$ then $f$ is a Lip(1) function and $g=-f$ by the \textit{c-transform} trick \cite{computationalOT}, and this reduces to 1-Wasserstein distance $W_1^d$. Consider the Kantorovich potential $f \defeq r$ as the reward function in \eqref{eq:reg_AL}, we conclude our proposition.

\qed

There are many ways of measuring distance between two probability measures, notably such as total variation, KL-divergence, $\chi^2$ distance and so on. None of them reflects the underlying geometric structure of the sample space, therefore the distance is sometimes ill-defined when the measures do not actually overlap \cite{improved_GAN}. On the other hand, Wasserstein distance, originated from optimal transport, is a canonical way of lifting geometry to define a proper distance metric between two distributions. Whereas the Kantorovich duality suggests the dual form of the OT as in \eqref{eq:ot_IRL}, it allows the application of stochastic gradient methods to make it eligible for large-scale problems. In a special case of Euclidean spaces $\set X= \set Y = \bb R^{|\SA|}$, by choosing $d(x, y) = \|y-x\|$, we can interprete the Wasserstein distance $W_1^d$ in \eqref{eq:claim} as follows,
\begin{align}
\sup_{r\sim \bb R^{\set S \times \set A}} \left\lbrace \bb E_{x\sim\bb R^{|\SA|}}\left[ r(x)(d\rho_E(x) - d\rho_\pi(x)) \right] \,\,\, :\,\,\, \|\nabla r\|_{\infty}\leq 1 \right\rbrace
\end{align}
The constraint on the gradient of reward function implies that the gradient norm at any point $x$ is upperbounded by $1$: $\|\nabla r\|_2 \leq 1$. This simple form suggests several ways of computing the Wasserstein distance by enforcing the Lipschitz condition, such as weight clipping \cite{wgan} and gradient penalty \cite{improved_GAN}.

\textbf{Relation to generative adversarial imitation learning. }
In canonical form of OT, we consider a ground cost $d$ (not neccessarily as a distance), the Kantorovich potentials can be chosen differently as seen in \eqref{eq:ot_IRL}, as long as the potentials ($f$, $g$) are Lipschitz regular: $f(x)+g(y)\leq d(x,y)$. Given a discriminator $D_{w}$ parameterized by $w$, and let $f(y)\defeq \log(1-D_{w}(y))$ and $g(x)\defeq \log(D_{w}(x))$, the OT problem becomes: 
\begin{align}
	& \sup_{w} \bb E_{\pi_E}\left[\log(1-D_{w}(y))\right] + \bb E_{\pi}\left[\log(D_{w}(x))\right] \\
\text{s.t. } \,\,\, \log(1&-D_{w}(y)) + \log(D_{w}(x)) \leq d(x, y), \,\,\, \forall x\times y \sim \rho_{\pi}\times \rho_E \nonumber
\end{align} 
It is then obvious, if the ground cost $d(\cdot,\cdot)$ is a non-negative constant, the constraint always holds. Therefore it recovers the objective of generative adversarial imitation learning (\textsc{Gail}) \cite{GAIL}. Note that by choosing a second Kantorovich potential $g$, it will eventually relax the proposition (\ref{prop:claim}) to take arbitrary functions $f$ and $g$, however it does no longer resemble the apprenticeship learning in \eqref{eq:reg_AL}. This implies that if $f\neq g$, the Kantorovich potentials might not resemble a valid reward function, and in Sec. \ref{sec:background} we emphasized the same conclusion for \gail\ as well. 


\textbf{From Wasserstein distance to IRL.} Moreover, we remark that the cost function regularizer defined in \eqref{eq:GAIL_IRL} actually induces a Wassetstein distance of the occupancy measures $\rho_\pi$ and $\rho_E$. An natural question would be, how the cost function regularizer $\psi$ in the context of \gail\ looks like for the Wassetstein distance $W_d$ and which type of \irl\ problem it actually solves. Note we use cost function instead of reward in order to make it comparable to \gail.
\begin{proposition}
	If the cost regularizer for generative adversarial imitation learning is chosen as
	\[
	  \psi(c) = \sup_{\rho \in \cB} \langle c, \rho - \rho_E \rangle - W_d(\rho, \rho_E) , \quad \forall c:\SA \to \R
	\]
	then the method coincides with our approach, i.e. $\psi^*(\rho_\pi - \rho_E) = W_d(\rho_\pi, \rho_E)$. In particular the inverse reinforcement learning part becomes
	\[
	  \IRL_\psi(\pi_E) = \argmax_{c:\SA \to \R} \Big(\inf_{\rho \in \cB} W_d(\rho, \rho_E) - \langle c, \rho \rangle \Big) + \Big(\min_{\pi \in \Pi} -H(\pi) + \langle c, \rho_\pi \rangle\Big)
	\]
	\label{prop:wasserstein_irl}
\end{proposition}
On one hand, the first term in  $\IRL_\psi$ estimates an occupancy measure $\rho$ to minimize the Wasserstein distance to the expert occupancy measure $\rho_E$, while it regularizes on its induced expected cost $\langle c, \rho \rangle$. On the other hand, the second term of  $\IRL_\psi$ finds an policy that minimizes the induced expected cost $\langle c, \rho_\pi \rangle$ while it maximizes the causal entropy of $\pi$. Therefore, the convex conjugate $\psi^*$ of the cost regularizer reduces to the Wasserstein distance, 
and the \irl\ problem couples the Wasserstein distance minimization and the entropy maximization. More details of the proof can be found in the supplementary material Sec.~\ref{supp:proof}. 

Deriving from apprenticeship learning we extend the choices of reward function by leveraging \textsc{Ipm}. In particular, we formulate the apprenticeship learning by 1-Wasserstein distance. Moreover, the Kantorovich potentials in OT give a more general set of functions they can represent, \ie Lipschitiz regular. We show that this corresponds to a certain type of \irl\ problem. Thus it enables a wide applications for imitation learning by choosing a large set of reward functions. Furthermore,  similar to \gail,  we can summarise an efficient iterative procedure such that the objective in \eqref{eq:ot_IRL} updates the dual potentials to improve the reward estimation, while for the next the policy is improved guided by the reward so as to generate expert-like state-actions. 

\section{Wasserstein Adversarial Imitation Learning}
Following the proposition \ref{prop:claim}, we propose a practical imitation learning algorithm based on 1-Wassersein distance, where we use a single Kantorovich potential $r$ to represent a valid reward function. Although optimal transport theory is well established, it is known for the computational difficulty. Recent advances in computational optimal transport enable large-scale applications using stochastic gradient methods and parallel computing on a modern GPU platform \cite{NIPS2013_4927, pmlr-v84-blondel18a}. Moreover, OT solvers are also extended to semidiscrete and continuous-continuous cases with arbitrary parameterization on dual variables and ground cost \cite{DBLP:conf/iclr/SeguyDFCRB18, Genevay:2016:SOL:3157382.3157482}. For a more complete reading of computational OT, we refer readers to the short book review \cite{computationalOT}.

\textbf{Regularized optimal transport. } For many problems, it is infeasible to directly solve the OT problem because it is not easy to enforce the Lipschitz continuity of the rewards. Therefore, we resort to the entropic regularization of OT and cast the problem to a single convex optimization problem.  The regularized OT has been studied thoroughly recently and shows nice properties of strong convexity as well as a smooth approximation of the original OT problem \cite{wilson1968use, pmlr-v84-blondel18a}. Let the ground cost $d$ be a distance metric defined on the state-action space, we compute the regularized dual form of OT as following, for simplicity we denote $W_d$ as being 1-Wasserstein distance.
\begin{align}
  W_d(\rho_\pi, \rho_E) = \sup_{r:\SA \to \R} \E_{y \sim\rho_E}\left[r(y)\right] - \E_{x\sim\rho_\pi}\left[r(x)\right] + \E_{(x, y) \sim \rho_\pi \times \rho_E}[\Omega_{d, \varepsilon}(r, x, y)] \label{eq:reg_ot}
\end{align}
where
\[
  \Omega_{d, \varepsilon}(r, x, y) =  
  \begin{cases} 
  -\varepsilon e^{\frac{1}{\varepsilon}(r(y) - r(x) - d(x, y))} & \text{Entropy regularization} \\
  -\frac{1}{4\varepsilon}(r(y) - r(x) - d(x, y))_+^2      & L_2 \text{-regularization} \nonumber
  \end{cases}
\]
regularizes the reward function in such a way that it decreases the objective if $r$ is not a Lipschitz(1) function. This decrease is larger for small $\varepsilon$ and we obtain the solution of the original OT problem in the limit $\varepsilon \to 0$ (See more in \cite{DBLP:journals/moc/ChizatPSV18} for the proof and further details).

%
%
%
%
%

Note that the expert demonstrations, \ie $y=\left\lbrace s, a \right\rbrace_{i=1\ldots L}$, is typically a finite set, while samples from the policy can be infinite. Without loss of generality, we consider both as generic continuous density measures, and parameterize the reward $r_w$ by a deep neural network with parameters $w$. For continuous-continuous OT optimization, we take a stochastic ascend step on a mini-batch of samples from both expert and policy.    

\textbf{Policy gradient. } Following the reward update, we take a policy gradient step by maximizing the expected reward while regularizing the policy causal entropy with a factor $\lambda$. 
\begin{align}
	\nabla_\theta \pi_\theta = \nabla_\theta \bb E_{\rho_{\pi_\theta}}\left\lbrack \hat{r}(s, a)\right\rbrack + \lambda\nabla_\theta H(\pi_\theta)
\end{align}  
Note that the estimated reward $\hat{r}$ from the previous step is treated as a fixed reward. Similar to \cite{GAIL}, for a finite mini-batch of state-actions, the gradient on the causal entropy term can be rewritten as $\sum_{(s,a) \in \SA}(\nabla_\theta\rho_{\pi_\theta}(s,a))(-\log\pi_{\theta_\text{old}}(a\cond s))$ , such that the empirical policy gradient becomes, 
\begin{align}
\nabla_\theta \pi_\theta = \bb \sum_{(s,a) \in \SA} \nabla_\theta\rho_{\pi_\theta}(s, a)\left( \hat{r}(s, a) - \lambda\log\pi_{\theta_{\text{old}}}(a\cond s) \right) \label{eq:policy_gradient}
\end{align}  
This is the standard policy gradient with the Kantorovich potential and negative log-likelihood on the policy as a fixed reward. It reads that the policy entropy favors those rewards with uncertain state-actions. However, our reward function from the OT problem is a valid reward function after training, whereas the surrogate reward of \gail\ becomes constant and is not useful anymore after convergence. Finally, we employ the trust region policy optimization (TRPO) \cite{pmlr-v37-schulman15} to update the policy by taking a KL-constrained natural gradient step. 

The algorithm is listed in \algref{alg:wail} named as Wasserstein Adversarial Imitation Learning (\wail), since it follows a basic style of adversarial training. On one hand, the objective in \eqref{eq:reg_ot} is increased by minimizing the Wasserstein distance. On the other hand, with the policy gradient in \eqref{eq:policy_gradient}, it is driven towards expert region to generate more expert-like state-actions, which will in turn decrease the objective in \eqref{eq:reg_ot}. For completeness, we derive a theorem to show that the \algref{alg:wail} actually converges to an optimal solution $(r^*, \pi^*)$. The proof of Theorem. \ref{thm:conv} is detailed in supplementary material Sec. \ref{supp:proof}.

\begin{theorem}\label{thm:conv}\
	Let $d\in\mathcal{F}_b(\mathcal{X}\times\mathcal{Y})$ bounded continuous functions on $\mathcal{X}\times\mathcal{Y}$ and let $\delta_k$ denotes the upper bound for the KL-constrained step at the $k$-th round of the \algref{alg:wail}. If $\sum_{k>1} \sqrt{\delta_k}<\infty$, then the \algref{alg:wail} returns $r^*$ and its corresponding optimal policy $\pi^*$.  
\end{theorem}

\begin{algorithm}
	\caption{Wasserstein adversarial imitation learning}\label{alg:wail}
	\begin{algorithmic}[1]

		\State \textbf{Input:} Expert trajectories $\tau_E$, initial policy weights $\theta^{(0)}$, initial reward function weights $w^{(0)}$ and metric $d$; learning rates $\alpha$, $\beta$; regularization factors $\varepsilon$, $\lambda$.
		\For{$k$ iterations}
		\State Sample $l_1$ state action pairs $X = (x_1, \dots, x_{l_1})$ from the environment with policy $\pi_{\theta^{(k)}}$.
		\State Sample $l_2$ state action pairs $Y = (y_1, \dots, y_{l_2})$ from the expert data $\tau_E$.
		\State Take an OT step to update the reward:
		\begin{align}
			w^{(k+1)} \leftarrow w^{(k)} + \alpha \nabla_w \sum_{x \in X, y \in Y} \left[r_{w^{(k)}}(y) - r_{w^{(k)}}(x) + 
		\Omega_{d, \varepsilon}(r_{w^{(k)}}, x, y)\right] \nonumber
		\end{align}
		\State Take a KL-constrained natural gradient step to update policy:
		\begin{align}
			\theta^{(k+1)} \leftarrow \theta^{(k)} + \beta \left(\nabla_\theta \hat\E_{\rho_{\pi_{\theta^{(k)}}}}[\hat r_{w^{(k)}}(x)]  - \lambda\nabla_\theta H(\pi_{\theta^{(k)}}) \right) \nonumber
		\end{align}
		\EndFor
		\Return Final policy $\pi^* = \pi_{\theta^{(k)}}$ and reward $r^* = r_{w^{(k)}}$.

	\end{algorithmic}
\end{algorithm}

\section{Experiments}
We evaluate \algref{alg:wail} on both classic and high dimensional continuous control tasks whose environments are provided with true reward functions in OpenAI's gym \cite{gym16}. Similary to \cite{GAIL}, the expert demonstrations are sampled from policies that are pretrained by TRPO with generalised advantage estimation \cite{Schulmanetal_ICLR2016} in different sizes (number of trajectories). Each expert trajectory contains about $50$ state-action pairs. The TRPO training setup and the expert performance are detailed in supplementary matrial Sec. \ref{supp:env}. 

\textbf{Baselines.} 
To evaluate the imitation performance of \algref{alg:wail}, we compare \wail\ with two baselines, \ie \gail\ and behavior cloning (\textsc{Bc}). The baseline \gail\ uses a discrminator over the state-action space and takes the logarithm as the cost function: $\log(D(s,a))$. In \cite{GAIL}, authors show superior imitation performance over other approaches such as Feature expectation matching (\textsc{Fem}) and Game-theoretic apprenticeship learning (\textsc{Gtal}). Behavior cloning, on the other hand, is a straightforward baseline to compare with, where no environment interaction is required but only expert demonstrations. All methods considered in our experiments do not assume the existence of the true reward function, but estimate the optimal policy directly. 

For all the methods, we parameterize the policy and value function with the same neural network architecture, which have two layers of $100$ units and $\text{tanh}$ activation. We model a stochastic policy, which outputs parameters of a normal distribution with a diagonal covariance. \bc\ is trained via maximizing the log-likelihood of the expert demonstrations with Adam \cite{adam14}. For the discriminator in \gail\ and the reward function in \textsc{Wail}, we use the same network architecture as policy network. In \textsc{Wail}, we choose the euclidean distance in the state-action space as the ground transport cost. Moreover, we found the $L_2$-regularization for OT is more stable and less hyperparameter sensitive, therefore we use $\Omega_{L_2}$ through out our experiments. The learning rate and regularization factors for all the three methods are fine-tuned to achieve optimal results. Finally, the parameters and environment interactions of the TRPO steps are the same as for the training of the expert policies. 

\begin{figure}[!ht]
	\subfloat[Classic control and MuJoCo environments\label{fig:exp_others}]{%
		\includegraphics[width=0.74\textwidth, valign=b]{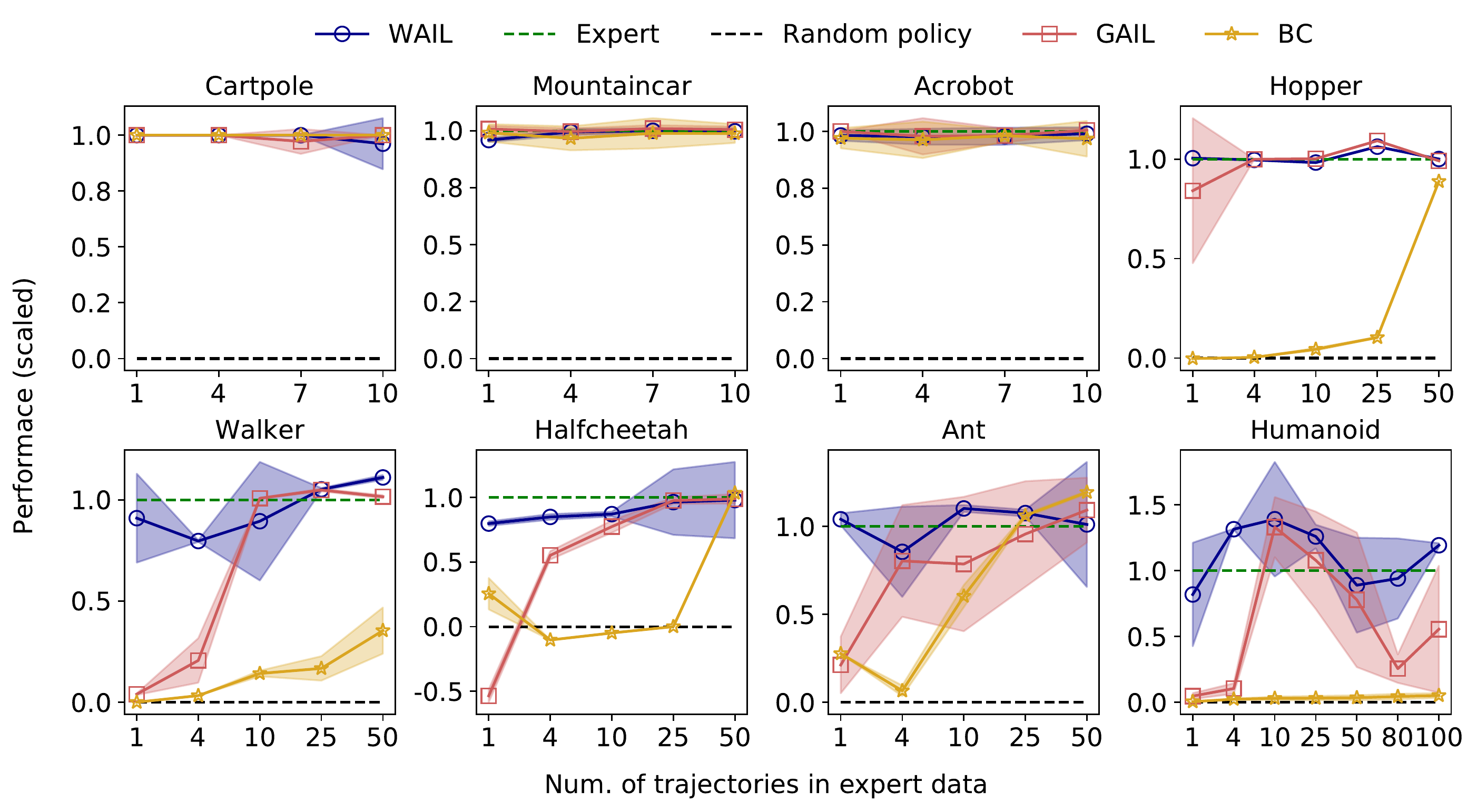}
	}
	\subfloat[Reacher\label{fig:exp_reacher}]{%
		\includegraphics[width=0.25\textwidth, valign=b]{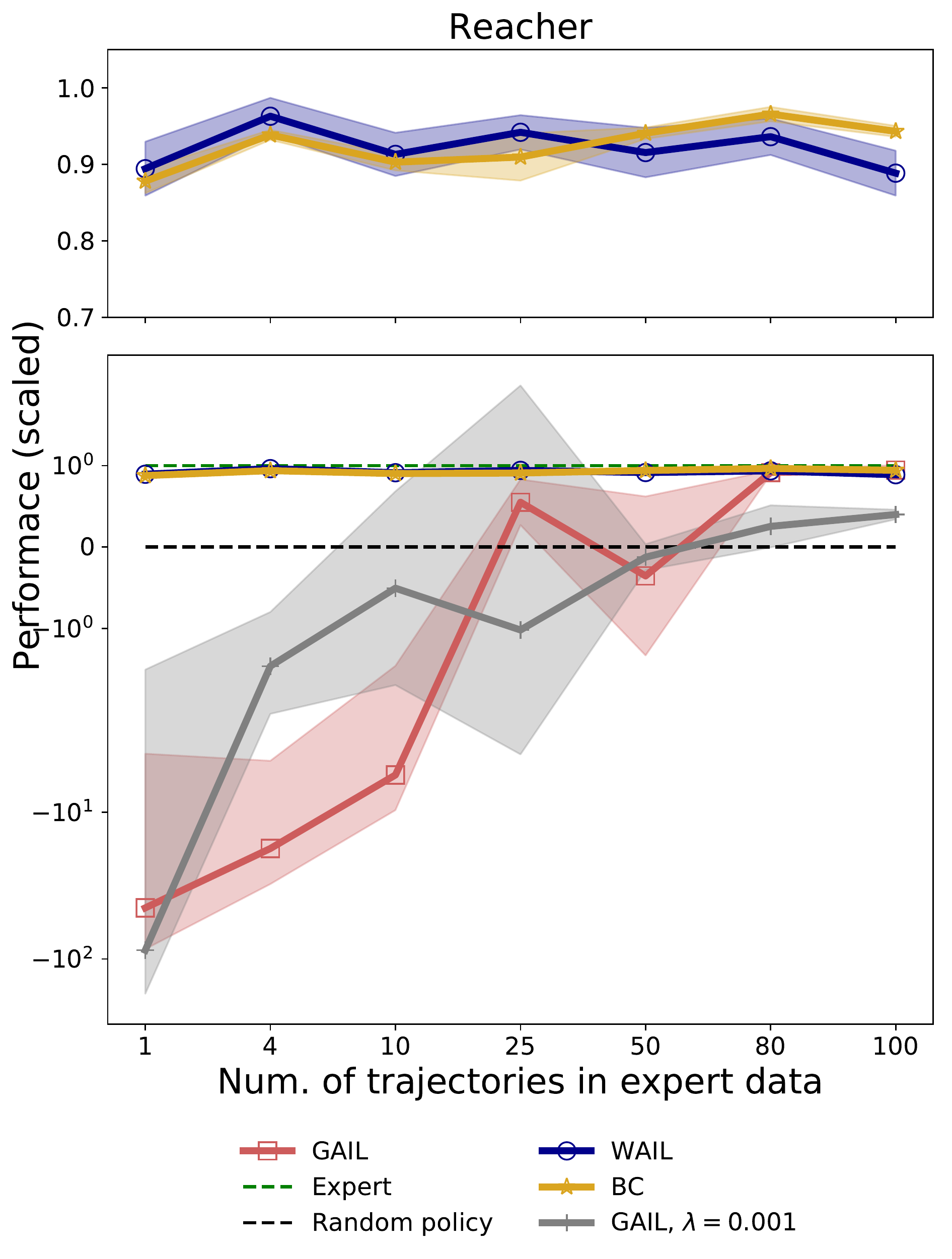}
	}
	\caption{Imitation performance of \textsc{Wail}, \textsc{Gail} and \textsc{Bc} on $9$ control tasks with respect to different expert data sizes. The performance is the average cumulative reward over $50$ trajectories and scaled in $\left\lbrack 0,1\right\rbrack$ with respect to expert and random policy performance.}
	\label{fig:exp_result}
\end{figure}

\textbf{Results. } We evaluate our appoach \wail\ on $9$ different control tasks in terms of expert sample complexity and show the result in \figref{fig:exp_result}. To compare with other baselines, we compute the averaged environment reward by randomly sampling $50$ trajectories using the policy learned and scale it by taking expert reward as $1.0$ and random policy reward as $0.0$. It shows that \wail\ outperforms both \gail\ and \bc\ on almost all the learning tasks with varying data sizes, and in particular we observe that \wail\ is extremely expert sample efficient. For all the tasks, only one expert trajectory is suffcient for \wail\ to approach expert behavior. In classic control tasks (Cartpole, Mountaincar and Acrobot), all the three methods achieve nearly-expert performance even with only one demonstration. For all the high-dimensional MuJoCo envrionments except Reacher, \bc\ can only imitate expert's behavior when trained with enough demonstrations, while \gail\ shows more promising results over \bc\ in terms of expert sample sizes. On the other hand, \wail\ dominates all the tasks in almost all the sample sizes, even though there is only one expert demonstration. Note that we observe a performance drop for the Humanoid task when we increase the data sizes, which might be incurred by the higher variance of expert policy. For the Reacher task, which is notably more difficult to learn, both \wail\ and \bc\ perform consistently well over varing data sizes (see the upper part of Fig.(\ref{fig:exp_reacher}) for the zoom-in view). However, \gail\ can only achieve expert's performance when we feed more than $80$ demonstrations. Advised by \cite{GAIL}, the causal entropy term improves \gail\ slightly only for small sample sizes. Finally, we report the full experiment results in supplementary material Sec. \ref{supp:exp}. 

\begin{figure}[!ht]
	\centering
	\includegraphics[width=\textwidth]{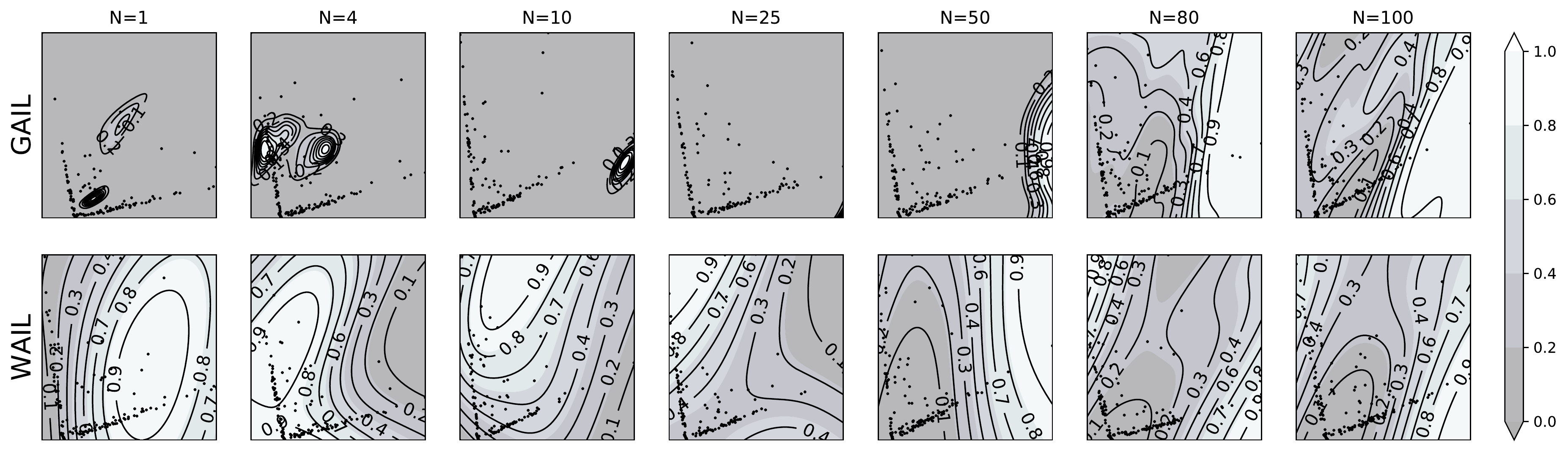}
	\caption{Reward surfaces of \textsc{Wail} and \textsc{Gail} on \textit{Humanoid} with respect to different expert data sizes.}
	\label{fig:reward_surface_humanoid}
\end{figure}

\textbf{Reward surface. }At Last, we demonstrate the reward surface of both \wail\ and \gail\ on the \textit{Humanoid} environment, which has the highest number of dimensions of states ($376$) and actions ($17$) among the experiments. The reward surface is drawn on a $2$-D board where the expert state-actions are projected onto via \textsc{Pca} \cite{Abdi2010PCA}. After training, we take the reward function: $r(s,a)$ for \wail\ and the negative Log-discriminator: $-\log(D(s,a))$ for \gail\ to compute the reward score for each discretized point on the $2$-D board, where each point is projected back to the state-action space via inverse \textsc{Pca}. The reward surface is then scaled to $[0, 1]$ and presented in \figref{fig:reward_surface_humanoid}. The projected state-actions spread in the lower-left region as two wings. For \wail, the learned reward function assigns higher reward along the directions of wings. When the data sizes is small, \eg $N=1$, rewards are concentrated on either wing. For $N \geq 80$ we see a smooth reward function with increasing rewards along both wings. On the other hand in \gail, for small datasets, the discriminator saturates and fails to imitate the experts. With increasing number of demonstrations, \eg $N=25$, the discriminator only learns to differentiate expert and others, eventually assigns constant reward almost everywhere. Thus it can not serve as a proper reward to update the policy afterwards. In summary,
the reward surface for \wail\ is obviously much smoother than \gail\ for all the expert data sizes. Although for a large portion of the $2$-D board, the reverse-projected state-actions might fall out of expert support, the reward scores computed via our approach is still well defined thanks to the geometric property of Wasserstein distance.

\section{Conclusion}
In this work, a natural connection between apprenticeship learning, optimal transport and \irl\ is built and justified theoretically. Upon the observation,  we present a novel imitation learning approach based on Wasserstein distance and enables the choices of smooth reward functions in the state-action space. Our approach is model free and outputs the dual function as intermediate result that can serve as a proper reward function. In several robotic control tasks, we demonstrate that our approach dominates other baselines by a large margin, in particular, it achieves expert behavior even with a extremely small number of expert demonstrations. This property might benefit from the optimization of OT problem and the smooth reward function. We leave the analysis of the sample complexity to the future work. 

Moreover, we remark that our approach \wail\ offers advantages in three folds. First, the ground cost can be enforced with a properly defined prior which encodes domain knowledge about the state-action space, or can even be estimated simutaneously. Second, the reward function is smooth and well defined. When it is parameterized by a neural network, it can represent a much more expressive family of reward function. Third, intrisically the optimal transport learns a transportation map in state-action space. It implies that the reward function can be transfered to different learning tasks as long as we can estimate such a tranport map. We will leave the justification of the remarks to the future work.


\bibliography{myrefs}

\begin{thebibliography}{10}

\bibitem{Abbeel2008ParkingLotNavigation}
P.~Abbeel, D.~Dolgov, A.~Y. Ng, and S.~Thrun.
\newblock Apprenticeship learning for motion planning with application to
  parking lot navigation.
\newblock pages 1083--1090, Sept 2008.

\bibitem{Abbeel:2004:ALV:1015330.1015430}
P.~Abbeel and A.~Y. Ng.
\newblock Apprenticeship learning via inverse reinforcement learning.
\newblock In {\em Proceedings of the Twenty-first International Conference on
  Machine Learning}, ICML '04, pages 1--, New York, NY, USA, 2004. ACM.

\bibitem{Abdi2010PCA}
H.~Abdi and L.~J. Williams.
\newblock Principal component analysis.
\newblock {\em WIREs Comput. Stat.}, 2(4):433--459, July 2010.

\bibitem{Argall2009SurveyLfD}
B.~D. Argall, S.~Chernova, M.~Veloso, and B.~Browning.
\newblock A survey of robot learning from demonstration.
\newblock {\em Robot. Auton. Syst.}, 57(5):469--483, May 2009.

\bibitem{wgan}
M.~Arjovsky, S.~Chintala, and L.~Bottou.
\newblock {W}asserstein generative adversarial networks.
\newblock In D.~Precup and Y.~W. Teh, editors, {\em Proceedings of the 34th
  International Conference on Machine Learning}, volume~70 of {\em Proceedings
  of Machine Learning Research}, pages 214--223, International Convention
  Centre, Sydney, Australia, 06--11 Aug 2017. PMLR.

\bibitem{Bain1999BehavioralCloning}
M.~Bain and C.~Sammut.
\newblock A framework for behavioural cloning.
\newblock In {\em Machine Intelligence 15, Intelligent Agents [St. Catherine's
  College, Oxford, July 1995]}, pages 103--129, Oxford, UK, UK, 1999. Oxford
  University.

\bibitem{pmlr-v84-blondel18a}
M.~Blondel, V.~Seguy, and A.~Rolet.
\newblock Smooth and sparse optimal transport.
\newblock In A.~Storkey and F.~Perez-Cruz, editors, {\em Proceedings of the
  Twenty-First International Conference on Artificial Intelligence and
  Statistics}, volume~84 of {\em Proceedings of Machine Learning Research},
  pages 880--889, Playa Blanca, Lanzarote, Canary Islands, 09--11 Apr 2018.
  PMLR.

\bibitem{gym16}
G.~Brockman, V.~Cheung, L.~Pettersson, J.~Schneider, J.~Schulman, J.~Tang, and
  W.~Zaremba.
\newblock Openai gym, 2016.
\newblock cite arxiv:1606.01540.

\bibitem{DBLP:journals/moc/ChizatPSV18}
L.~Chizat, G.~Peyr{\'{e}}, B.~Schmitzer, and F.~Vialard.
\newblock Scaling algorithms for unbalanced optimal transport problems.
\newblock {\em Math. Comput.}, 87(314):2563--2609, 2018.

\bibitem{NIPS2013_4927}
M.~Cuturi.
\newblock Sinkhorn distances: Lightspeed computation of optimal transport.
\newblock In C.~J.~C. Burges, L.~Bottou, M.~Welling, Z.~Ghahramani, and K.~Q.
  Weinberger, editors, {\em Advances in Neural Information Processing Systems
  26}, pages 2292--2300. Curran Associates, Inc., 2013.

\bibitem{Finn2016GuidedCostLearning}
C.~Finn, S.~Levine, and P.~Abbeel.
\newblock Guided cost learning: Deep inverse optimal control via policy
  optimization.
\newblock In {\em Proceedings of the 33nd International Conference on Machine
  Learning, {ICML} 2016, New York City, NY, USA, June 19-24, 2016}, pages
  49--58, 2016.

\bibitem{fu2018learning}
J.~Fu, K.~Luo, and S.~Levine.
\newblock Learning robust rewards with adverserial inverse reinforcement
  learning.
\newblock In {\em International Conference on Learning Representations}, 2018.

\bibitem{Genevay:2016:SOL:3157382.3157482}
A.~Genevay, M.~Cuturi, G.~Peyr{\'e}, and F.~Bach.
\newblock Stochastic optimization for large-scale optimal transport.
\newblock In {\em Proceedings of the 30th International Conference on Neural
  Information Processing Systems}, NIPS'16, pages 3440--3448, USA, 2016. Curran
  Associates Inc.

\bibitem{Goodfellow:2014:GAN}
I.~J. Goodfellow, J.~Pouget-Abadie, M.~Mirza, B.~Xu, D.~Warde-Farley, S.~Ozair,
  A.~Courville, and Y.~Bengio.
\newblock Generative adversarial nets.
\newblock In {\em Proceedings of the 27th International Conference on Neural
  Information Processing Systems - Volume 2}, NIPS'14, pages 2672--2680,
  Cambridge, MA, USA, 2014. MIT Press.

\bibitem{improved_GAN}
I.~Gulrajani, F.~Ahmed, M.~Arjovsky, V.~Dumoulin, and A.~C. Courville.
\newblock Improved training of wasserstein gans.
\newblock In I.~Guyon, U.~V. Luxburg, S.~Bengio, H.~Wallach, R.~Fergus,
  S.~Vishwanathan, and R.~Garnett, editors, {\em Advances in Neural Information
  Processing Systems 30}, pages 5767--5777. Curran Associates, Inc., 2017.

\bibitem{GAIL}
J.~Ho and S.~Ermon.
\newblock Generative adversarial imitation learning.
\newblock In D.~D. Lee, M.~Sugiyama, U.~V. Luxburg, I.~Guyon, and R.~Garnett,
  editors, {\em Advances in Neural Information Processing Systems 29}, pages
  4565--4573. Curran Associates, Inc., 2016.

\bibitem{Ho:2016:MIL:3045390.3045681}
J.~Ho, J.~K. Gupta, and S.~Ermon.
\newblock Model-free imitation learning with policy optimization.
\newblock In {\em Proceedings of the 33rd International Conference on
  International Conference on Machine Learning - Volume 48}, ICML'16, pages
  2760--2769. JMLR.org, 2016.

\bibitem{adam14}
D.~Kingma and J.~Ba.
\newblock Adam: A method for stochastic optimization.
\newblock {\em International Conference on Learning Representations}, 12 2014.

\bibitem{Kuderer2013TeachingMobileRobots}
M.~Kuderer, H.~Kretzschmar, and W.~Burgard.
\newblock Teaching mobile robots to cooperatively navigate in populated
  environments.
\newblock In {\em Proc.~of the IEEE/RSJ International Conference on Intelligent
  Robots and Systems (IROS)}, Tokyo, Japan, 2013.

\bibitem{levine2010feature}
S.~Levine, Z.~Popovic, and V.~Koltun.
\newblock Feature construction for inverse reinforcement learning.
\newblock In {\em Advances in Neural Information Processing Systems}, pages
  1342--1350, 2010.

\bibitem{Levine2011GPIRL}
S.~Levine, Z.~Popovic, and V.~Koltun.
\newblock Nonlinear inverse reinforcement learning with gaussian processes.
\newblock In J.~Shawe-Taylor, R.~S. Zemel, P.~L. Bartlett, F.~Pereira, and
  K.~Q. Weinberger, editors, {\em Advances in Neural Information Processing
  Systems 24}, pages 19--27. Curran Associates, Inc., 2011.

\bibitem{Ng2000AlgorithmsForIRL}
A.~Y. Ng and S.~J. Russell.
\newblock Algorithms for inverse reinforcement learning.
\newblock In {\em Proceedings of the Seventeenth International Conference on
  Machine Learning}, ICML '00, pages 663--670, San Francisco, CA, USA, 2000.
  Morgan Kaufmann Publishers Inc.

\bibitem{computationalOT}
G.~Peyré and M.~Cuturi.
\newblock Computational optimal transport.
\newblock {\em Foundations and Trends in Machine Learning}, 11 (5-6):355--602,
  2019.

\bibitem{ERILross10a}
S.~Ross and D.~Bagnell.
\newblock Efficient reductions for imitation learning.
\newblock In Y.~W. Teh and M.~Titterington, editors, {\em Proceedings of the
  Thirteenth International Conference on Artificial Intelligence and
  Statistics}, volume~9 of {\em Proceedings of Machine Learning Research},
  pages 661--668, Chia Laguna Resort, Sardinia, Italy, 13--15 May 2010. PMLR.

\bibitem{Russell1998LearningAgents}
S.~Russell.
\newblock Learning agents for uncertain environments.
\newblock In {\em Proceedings of the eleventh annual conference on
  Computational learning theory}, pages 101--103. ACM, 1998.

\bibitem{pmlr-v37-schulman15}
J.~Schulman, S.~Levine, P.~Abbeel, M.~Jordan, and P.~Moritz.
\newblock Trust region policy optimization.
\newblock In F.~Bach and D.~Blei, editors, {\em Proceedings of the 32nd
  International Conference on Machine Learning}, volume~37 of {\em Proceedings
  of Machine Learning Research}, pages 1889--1897, Lille, France, 07--09 Jul
  2015. PMLR.

\bibitem{Schulmanetal_ICLR2016}
J.~Schulman, P.~Moritz, S.~Levine, M.~Jordan, and P.~Abbeel.
\newblock High-dimensional continuous control using generalized advantage
  estimation.
\newblock In {\em Proceedings of the International Conference on Learning
  Representations (ICLR)}, 2016.

\bibitem{DBLP:conf/iclr/SeguyDFCRB18}
V.~Seguy, B.~B. Damodaran, R.~Flamary, N.~Courty, A.~Rolet, and M.~Blondel.
\newblock Large scale optimal transport and mapping estimation.
\newblock In {\em {ICLR}}. OpenReview.net, 2018.

\bibitem{sriperumbudur2012empirical}
B.~K. Sriperumbudur, K.~Fukumizu, A.~Gretton, B.~Sch{\"o}lkopf, G.~R.
  Lanckriet, et~al.
\newblock On the empirical estimation of integral probability metrics.
\newblock {\em Electronic Journal of Statistics}, 6:1550--1599, 2012.

\bibitem{Syed:2008:ALU:1390156.1390286}
U.~Syed, M.~Bowling, and R.~E. Schapire.
\newblock Apprenticeship learning using linear programming.
\newblock In {\em Proceedings of the 25th International Conference on Machine
  Learning}, ICML '08, pages 1032--1039, New York, NY, USA, 2008. ACM.

\bibitem{Syed2007GameTheoreticAL}
U.~Syed and R.~E. Schapire.
\newblock A game-theoretic approach to apprenticeship learning.
\newblock In J.~C. Platt, D.~Koller, Y.~Singer, and S.~T. Roweis, editors, {\em
  NIPS}, pages 1449--1456. Curran Associates, Inc., 2007.

\bibitem{tail2018socially}
L.~Tail, J.~Zhang, M.~Liu, and W.~Burgard.
\newblock Socially compliant navigation through raw depth inputs with
  generative adversarial imitation learning.
\newblock In {\em 2018 IEEE International Conference on Robotics and Automation
  (ICRA)}, pages 1111--1117. IEEE, 2018.

\bibitem{Wang_WIOC}
Y.~Wang, L.~Song, and H.~Zha.
\newblock Learning to optimize via wasserstein deep inverse optimal control.
\newblock {\em CoRR}, abs/1805.08395, 2018.

\bibitem{wilson1968use}
A.~Wilson.
\newblock {\em The Use of Entropy Maximising Models in the Theory of Trip
  Distribution, Mode Split and Route Split}.
\newblock Working papers // Centre for Environmental Studies. Centre for
  Environmental Studies, 1968.

\bibitem{ZhouMCEIRL}
Z.~{Zhou}, M.~{Bloem}, and N.~{Bambos}.
\newblock Infinite time horizon maximum causal entropy inverse reinforcement
  learning.
\newblock {\em IEEE Transactions on Automatic Control}, 63(9):2787--2802, Sep.
  2018.

\bibitem{Ziebart2010Modeling}
B.~D. Ziebart, J.~A. Bagnell, and A.~K. Dey.
\newblock Modeling interaction via the principle of maximum causal entropy.
\newblock In {\em Proc. of the International Conference on Machine Learning},
  pages 1255--1262, 2010.

\bibitem{Ziebart2008MaxEntIRL}
B.~D. Ziebart, A.~Maas, J.~A.~D. Bagnell, and A.~Dey.
\newblock Maximum entropy inverse reinforcement learning.
\newblock In {\em Proceeding of AAAI 2008}, July 2008.

\end{thebibliography}
\bibliographystyle{abbrv}

\clearpage
\appendix

\section{Proofs}
\label{supp:proof}

\subsection{Proof to Proposition \ref{prop:wasserstein_irl}}
\label{supp:proof1}
Let $c = -r$ be the corresponding cost function for the reward $r$. Let $\bar{\cB} := \{\rho - \rho_E \mid \rho \in \cB\}$. If we consider the function $f:\cB \to \R$ defined by $f(\rho) = W(\rho + \rho_E, \rho_E)$
then we can see that it is convex and lower semi-continuous as a pointwise supremum of linear functions. Hence $f^{**} = f$. Moreover we have that
\begin{align*}
f^*(c) & = \sup_{\rho \in \bar{\cB}} \langle c, \rho \rangle - W(\rho + \rho_E, \rho_E) 
= \sup_{\rho \in \cB} \langle c, \rho - \rho_E \rangle - W(\rho, \rho_E) 
= \psi(c)
\end{align*}
and hence $\psi^*(\rho_\pi - \rho_E) = f(\rho_\pi - \rho_E) = W(\rho_\pi, \rho_E)$. The second claim follows by pluging in $\psi$ into the \textsc{Irl} formula from \cite{GAIL}
\[
\IRL_\psi(\pi_E) = \argmax_{c:\SA \to \R} -\psi(c) + \Big(\min_{\pi \in \Pi} -H(\pi) + \langle c, \rho_\pi \rangle\Big) - \langle c, \rho_E \rangle.
\]
\qed

\subsection{Proof to Theorem \ref{thm:conv}}
\label{supp:proof2}
Let us denote the function on the right hand side of \eqref{eq:reg_ot}, by $L_w(\theta)$, i.e., 
\begin{align}
&L_w(\theta):= \E_{y \sim\rho_E}\left[r(y)\right] - \E_{x\sim\rho_\pi}\left[r(x)\right] + \E_{(x, y) \sim \rho_\pi \times \rho_E}[\Omega_{d, \varepsilon}(r, x, y)]\nonumber
\end{align}
where $w$ and $\theta$ parameterize the reward function and the policy. First, we show that for any given $w$, $L_w(\theta^{(k)})$ in which $\{\theta^{(k)}\}$ are obtained from the policy gradient step in \algref{alg:wail} converges. To do so, it suffices to show that $\{L_w(\theta^{(k)})\}$ is Cauchy. 
\begin{align*}
&|L_w(\theta^{(k)})- L_w(\theta^{(k+m)})|\leq |L_w(\theta^{(k)})- L_w(\theta^{(k+1)})| + \cdots +  |L_w(\theta^{(k+m-1)})- L_w(\theta^{(k+m)})|\\
&\leq 2M\left(d_{TV}(\pi_{\theta^{(k)}},\pi_{\theta^{(k+1)}})+ \cdots + d_{TV}(\pi_{\theta^{(k+m-1)}},\pi_{\theta^{(k+m)}})\right)\leq M \sum_{i=k}^{k+m}\sqrt{2\delta_i},
\end{align*}
where $d_{TV}(\cdot, \cdot)$ denotes the total variation distance. In the above derivation, we also use the fact that $|r_w(y) - r_w(x) + \Omega_{c, \varepsilon}(r_w(x), r_w(y))|\leq M$ for any $w$ around the $\sup_w L_w(\theta)$.
The second inequality is due to the definition of total variation. 
The last inequality is because of the Pinsker's inequality and the fact that based on the policy gradient update, the KL-divergence between two consecutive policies is bounded by $\delta$s. The above inequality and the assumption in Theorem \ref{thm:conv} imply that $\{L_w(\theta^{(k)})\}$ is Cauchy and hence it converges. Furthermore, since the convergence is independent of $w$, it converges uniformly. 

On the other hand, since for fixed parameter $\theta$, $L_w(\theta)$ is a continuous and concave function of $w$, there exists $\theta^*$ such that $\sup_w L_{w}(\theta^{(k)})\rightarrow\sup_w L_{w}(\theta^*)$, when $L_{w}(\theta^{(k)}$ converges uniformly to $L_{w}(\theta^*)$. 
The concavity implies that $w^*_k:=\arg\sup_wL_w(\theta^{(k)})$ also converges to $w^* :=\arg\sup_wL_w(\theta^*)$. 
Moreover, assuming $\nabla_w L_w(\theta^{(k)})$ is uniformly continuous with respect to $w$, we obtain that $\nabla_w L_w(\theta^{(k)})\rightarrow \nabla_w L_w(\theta)$.
These results imply that $w^{(k)}$ obtained from the gradient ascend step in \algref{alg:wail} will converge to $w^*$ and concludes the Theorem.

\qed

\section{Environments and expert policies}
\label{supp:env}
We run experiments on expert demonstrations trained by TRPO with generalised advantage estimation on 9 different control tasks, the environments details and performance of both expert and random policies are listed in \tblref{tbl:trpo_environment}. Note expert performance is evaluated by sampling $50$ trajectories from the pretrained expert policy, while random policy is initialized randomly and sampled for $50$ trajectories to compute random policy performance.

\begin{table}[htb!]
	\centering
	\caption{Environments details and performance of both random and expert policies.}
	\label{tbl:trpo_environment}
	\scalebox{0.86}{
		\begin{tabular}{lcccc}
			\toprule
			Env. name & State dimension & Action dimension & Random policy performance & Expert performance \\ 
			\midrule 
			CartPole-v1 & $4$ & $2$ & $23.10 \pm 12.69$ & $500.00 \pm 0.00$ \\
			Acrobot-v1 & $6$ & $3$ & $-494.72 \pm 30.21$ & $-69.04 \pm 5.06$ \\
			MountainCar-v0 & $2$ & $3$ & $-500.00 \pm 0.00$ & $-98.38 \pm 7.63$ \\
			\midrule
			Hopper-v2 & $11$ & $3$ & $17.68 \pm 12.17$ & $7089.25 \pm 6.55$ \\
			Walker2d-v2 & $17$ & $6$ & $1.04 \pm 6.33$ & $8281.99 \pm 102.76$ \\
			HalfCheetah-v2 & $17$ & $6$ & $-1321.26 \pm 85.45$ & $8777.48 \pm 135.39$ \\
			Ant-v2 & $111$ & $8$ & $-489.26 \pm 1392.61$ & $7580.65 \pm 1279.01$ \\
			Humanoid-v2 & $376$ & $17$ & $103.72 \pm 34.73$ & $8266.71 \pm 3340.75$ \\
			Reacher-v2 & $11$ & $2$ & $-4370.74 \pm 128.50$ & $-231.18 \pm 214.45$ \\
			\bottomrule
		\end{tabular}
	}
\end{table}

\begin{table}[htb!]
	\centering
	\caption{TRPO training parameters and expert performance}
	\label{tbl:trpo_parameters}
	\scalebox{0.92}{
		\begin{tabular}{lccccccc}
			\toprule
			Env. name & Batch size & Iterations & $\gamma$ & $\tau$ & max. episode steps & max. KL & Damping \\ 
			\midrule 
			CartPole-v1 & $20000$ & $200$ & $0.99$ & $0.99$ & $500$ & $0.01$ & $0.01$\\
			Acrobot-v1 & $20000$ & $200$ & $0.99$ & $0.99$ &  $500$ & $0.01$ & $0.01$ \\
			MountainCar-v0 & $20000$ & $200$ & $0.99$ & $0.99$ & $500$ & $0.01$ & $0.01$ \\
			\midrule
			Hopper-v2  & $50000$ & $500$ & $0.995$ & $0.99$ & $2000$ & $0.01$ & $0.01$ \\ 
			Walker2d-v2 & $50000$ & $500$ & $0.995$ & $0.99$ & $2000$ & $0.01$ & $0.01$ \\ 
			HalfCheetah-v2 & $50000$ & $500$ & $0.995$ & $0.97$ & $2000$ & $0.01$ & $0.01$ \\
			Ant-v2 & $50000$ & $500$ & $0.995$ & $0.97$ & $2000$ & $0.01$ & $0.01$ \\
			Humanoid-v2 & $50000$ & $1500$ & $0.995$ & $0.97$ & $2000$ & $0.01$ & $0.01$ \\ 
			Reacher-v2 & $50000$ & $200$ & $0.99$ & $0.95$ & $2000$ & $0.01$ & $0.01$\\
			\bottomrule
		\end{tabular}
	}
\end{table}

In \tblref{tbl:trpo_parameters}, we list the TRPO training parameters for the expert policies, where the parameters $\gamma$ and $\tau$ are two discount factors trading off the bias-variance
and preserve the same meaning as stated in \cite{Schulmanetal_ICLR2016}. 
For classic control tasks, \ie CartPole, MountainCar and Acrobot, we chose the maximal episode steps as $500$, while for the other tasks they are allowed to be rolled out for longer trajectories with $2000$ time steps.

\section{Experiment details and further results}
\label{supp:exp}
We first describe the training parameters for our approach \wail. In \tblref{tbl:wail_parameters}, the ground cost and type of regularization are fixed through out all experiments. Regularization factor $\varepsilon$ and learning rate $\alpha$ are fine-tuned to achieve optimal performance for each task. We train \wail with the same amount of environment interactions as how we train the TRPO policies for experts, however the task \textit{Ant} requires more iterations to converge.  

\begin{figure}[!ht]
	\includegraphics[width=\textwidth]{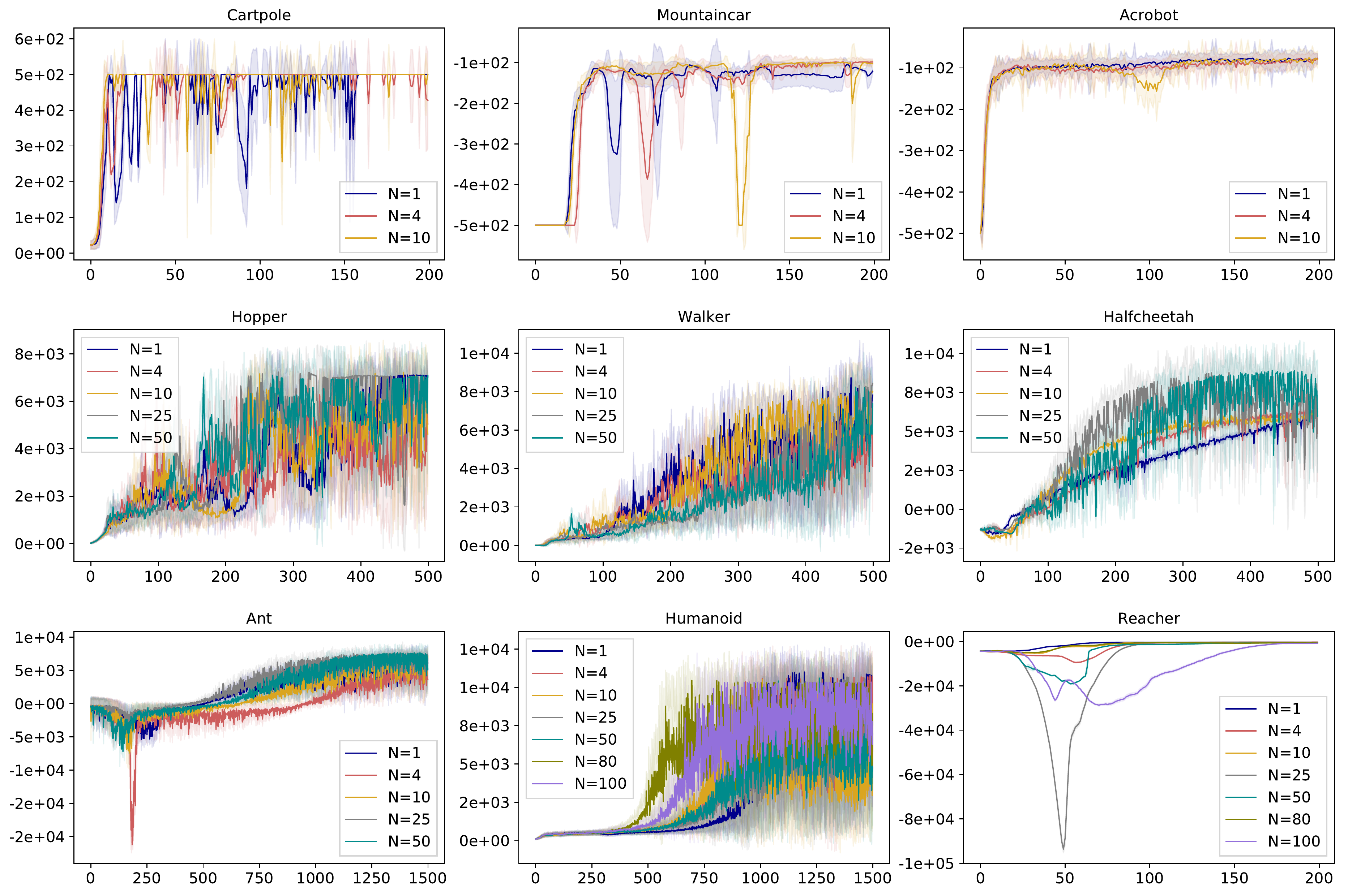}
	\caption{Training curves of \textsc{Wail} for all control tasks with respect to different expert data sizes.}
	\label{fig:exp_training_curve}
\end{figure}

\begin{table}[htb!]
	\centering
	\caption{Training parameters for \textsc{Wail}}
	\label{tbl:wail_parameters}
	\scalebox{1.0}{
		\begin{tabular}{lccccc}
			\toprule
			Env. name & Iterations & Reg. value $\varepsilon$ & Learning rate $\alpha$ &  Ground cost & Regularization\\ 
			\midrule 
			CartPole-v1 & $200$ & $0.01$ & $0.01$ & Euclidean & $L_2$\\
			Acrobot-v1 & $200$ & $0.001$ & $0.01$ & Euclidean& $L_2$\\
			MountainCar-v0 & $200$ & $0.001$ & $0.01$ & Euclidean& $L_2$\\
			\midrule
			Hopper-v2  & $500$ & $0.01$ & $0.01$ & Euclidean& $L_2$\\ 
			Walker2d-v2 & $500$ & $0.01$ & $0.01$ & Euclidean& $L_2$\\ 
			HalfCheetah-v2 & $500$ & $0.001$ & $0.01$ & Euclidean& $L_2$\\ 
			Ant-v2 & $1500$ & $0.0	1$ & $0.01$ & Euclidean& $L_2$\\  
			Humanoid-v2 & $1500$ & $0.001$ & $0.01$ & Euclidean& $L_2$\\ 
			Reacher-v2 & $200$ & $0.001$ & $0.01$ & Euclidean& $L_2$\\
			\bottomrule
		\end{tabular}
	}
\end{table}

In the paper, we complement the experiment results in \tblref{tbl:fullexp} and \tblref{tbl:fullexp_reacher}, where we list all the results on the control tasks with respect to varying sizes of expert demonstrations. The performance is computed by averaging the rewards of the randomly sampled $50$ trajectories using the policy after training. For the task \textit{Reacher}, we also report results for \gail with additional causal entropy term. All the other tasks do not require the causal entropy term. 

\begin{table}[htb!]
	\centering
	\caption{Experiment results for \textsc{Bc},  \textsc{Gail} and \textsc{Wail} on robotic control tasks.}
	\label{tbl:fullexp}
	\scalebox{1.0}{
		\begin{tabular}{lcccc}
			\toprule
			Tasks & Dataset Sizes & \textsc{Bc}  &  \textsc{Gail} & \textsc{Wail}\\ 
			\midrule 
			CartPole-v1 & 1 & $500.00 \pm 0.00$ & $500.00 \pm 0.00$ & $500.00 \pm 0.00$ \\
			& 4 & $500.00 \pm 0.00$ & $500.00 \pm 0.00$ & $500.00 \pm 0.00$ \\
			& 7 & $500.00 \pm 0.00$ & $486.70 \pm 26.60$ & $500.00 \pm 0.00$ \\
			& 10 & $500.00 \pm 0.00$ & $500.00 \pm 0.00$ & $481.80 \pm 54.60$ \\
			MountainCar-v0 & 1 & $-106.81 \pm 15.22$ & $-100.10 \pm 5.59$ & $-119.40 \pm 3.41$ \\
			& 4 & $-111.52 \pm 21.20$ & $-99.70 \pm 5.66$ & $-100.20 \pm 5.56$ \\
			& 7 & $-108.29 \pm 26.33$ & $-100.00 \pm 5.87$ & $-103.90 \pm 4.04$ \\
			& 10 & $-107.97 \pm 16.37$ & $-101.20 \pm 5.36$ & $-104.10 \pm 4.83$ \\
			Acrobot-v1 & 1 & $-83.60 \pm 18.82$ & $-70.90 \pm 0.70$ & $-78.50 \pm 9.80$ \\
			& 4 & $-87.39 \pm 33.99$ & $-80.80 \pm 33.79$ & $-81.60 \pm 14.91$ \\
			& 7 & $-76.02 \pm 11.38$ & $-75.70 \pm 13.16$ & $-77.00 \pm 16.19$ \\
			& 10 & $-85.26 \pm 33.17$ & $-69.50 \pm 3.98$ & $-75.90 \pm 12.03$ \\
			\midrule
			Hopper-v2 & 1 & $-0.02 \pm 0.34$ & $5939.63 \pm 2567.72$ & $7091.54 \pm 4.43$ \\
			& 4 & $42.20 \pm 2.14$ & $7076.64 \pm 7.21$ & $7051.52 \pm 12.98$ \\
			& 10 & $329.95 \pm 43.55$ & $7086.81 \pm 5.40$ & $6951.40 \pm 19.22$ \\
			& 25 & $702.33 \pm 16.46$ & $7249.78 \pm 5.36$ & $7052.91 \pm 1.17$ \\
			& 50 & $6296.17 \pm 17.00$ & $7027.15 \pm 5.26$ & $7092.33 \pm 1.83$ \\
			Walker2d-v2 & 1 & $4.03 \pm 4.18$ & $354.67 \pm 35.16$ & $8032.39 \pm 1939.86$ \\
			& 4 & $327.36 \pm 4.10$ & $2043.18 \pm 1086.51$ & $7871.23 \pm 42.50$ \\
			& 10 & $1187.28 \pm 119.40$ & $8399.34 \pm 15.87$ & $7454.98 \pm 2441.02$ \\
			& 25 & $1347.15 \pm 484.91$ & $8432.00 \pm 54.68$ & $8457.88 \pm 52.53$ \\
			& 50 & $2934.55 \pm 945.70$ & $8410.72 \pm 63.66$ & $9199.59 \pm 97.23$ \\
			HalfCheetah-v2 & 1 & $1220.38 \pm 1212.46$ & $-6647.73 \pm 500.73$ & $6622.88 \pm 172.69$ \\
			& 4 & $-2346.06 \pm 0.09$ & $4178.90 \pm 325.69$ & $7138.54 \pm 217.60$ \\
			& 10 & $-1813.94 \pm 1.73$ & $6513.05 \pm 475.13$ & $7494.07 \pm 189.64$ \\
			& 25 & $-1315.82 \pm 22.05$ & $8538.20 \pm 255.30$ & $8417.88 \pm 2561.95$ \\
			& 50 & $9118.27 \pm 143.82$ & $8660.19 \pm 380.25$ & $8567.99 \pm 2982.91$ \\
			Ant-v2 & 1 & $1916.23 \pm 1.86$ & $1354.49 \pm 1398.33$ & $8551.13 \pm 296.60$ \\
			& 4 & $71.39 \pm 260.20$ & $6382.30 \pm 2714.80$ & $6825.33 \pm 2194.04$ \\
			& 10 & $4494.25 \pm 524.08$ & $6003.85 \pm 3156.21$ & $8612.25 \pm 161.19$ \\
			& 25 & $8588.86 \pm 85.31$ & $7689.52 \pm 2568.24$ & $8709.75 \pm 162.92$ \\
			& 50 & $9149.67 \pm 62.40$ & $8332.24 \pm 1487.32$ & $7672.79 \pm 2869.48$ \\
			Humanoid-v2 & 1 & $155.16 \pm 14.79$ & $597.32 \pm 233.80$ & $8625.99 \pm 4091.31$ \\
			& 4 & $295.51 \pm 39.02$ & $957.80 \pm 327.40$ & $10834.24 \pm 31.45$ \\
			& 10 & $313.15 \pm 95.03$ & $9293.30 \pm 1576.69$ & $9706.73 \pm 3004.75$ \\
			& 25 & $357.84 \pm 134.90$ & $8754.26 \pm 2968.61$ & $10205.11 \pm 718.40$ \\
			& 50 & $388.05 \pm 156.50$ & $6456.35 \pm 4170.96$ & $7363.81 \pm 2943.62$ \\
			& 80 & $531.73 \pm 201.71$ & $2579.94 \pm 1045.17$ & $9208.88 \pm 2944.76$ \\
			& 100 & $543.90 \pm 154.29$ & $4814.52 \pm 4092.85$ & $10222.15 \pm 135.78$ \\
			\bottomrule
		\end{tabular}
	}
\end{table}
\begin{table}[htb!]
	\centering
	\caption{Experiment results on Reacher-v2.}
	\label{tbl:fullexp_reacher}
	\scalebox{.87}{
		\begin{tabular}{ccccc}
			\toprule
			Dataset Sizes & \textsc{Bc} & \textsc{Gail}  &  \textsc{Wail} & \textsc{Gail}, $\lambda=0.001$\\ 
			\midrule 
			
			1 & $-559.91 \pm 68.47$ & $-198916.81 \pm 177278.54$ & $-489.25 \pm 153.27$ & $-382769.66 \pm 371866.05$  \\
			4 & $-463.15 \pm 27.30$ & $-77809.13 \pm 54919.16$ & $-361.42 \pm 100.12$ & $-10458.54 \pm 2768.60$ \\
			10 & $-535.07 \pm 45.80$ & $-27978.31 \pm 17410.60$ & $-491.86 \pm 120.21$ & $-6516.11 \pm 5052.92$ \\
			25 & $-561.73 \pm 129.18$ & $-2065.02 \pm 1172.93$ & $-426.11 \pm 93.96$ & $-8635.25 \pm 12574.90$ \\
			50 & $-476.51 \pm 30.87$ & $-5838.25 \pm 4041.17$ & $-581.27 \pm 134.10$ & $-4884.68 \pm 661.20$ \\
			80 & $-386.03 \pm 39.01$ & $-591.78 \pm 98.99$ & $-507.84 \pm 98.63$ & $-3321.55 \pm 1066.91$ \\
			100 & $-469.58 \pm 29.84$ & $-459.73 \pm 257.68$ & $-695.68 \pm 121.63$ & $-2717.09 \pm 244.47$ \\
			\bottomrule
		\end{tabular}
	}
\end{table}

Moreover, we include the training curves of all the tasks for \textsc{Wail} with respect to different dataset sizes. In \figref{fig:exp_training_curve}, the reward is computed over all sampled trajectories at each iteration during training, where the sampled trajectories are used for TRPO step as well.  

\section{Reward surface}
\label{supp:reward}
To verify our claim that the reward function learned from \wail is smoother and can be used as real reward, we show reward surface of both \wail and \gail on a $2$-D board. For all the tasks, we take the learned the reward function in \wail or negative cost function in \gail to compute the reward score of each point on the $2$-D surface, to where we project a subset of the expert samples (state-actions) using PCA. Note that each point on the $2$-D surface is transformed back to its state-action space and then is valued for its reward. In Fig.(2), we list all the reward surfaces for all tasks varing the expert dataset sizes. 

\begin{figure}[!ht]
	\centering
	\subfloat[CartPole\label{fig:reward_surface_cartpole}]{%
		\includegraphics[width=0.8\textwidth]{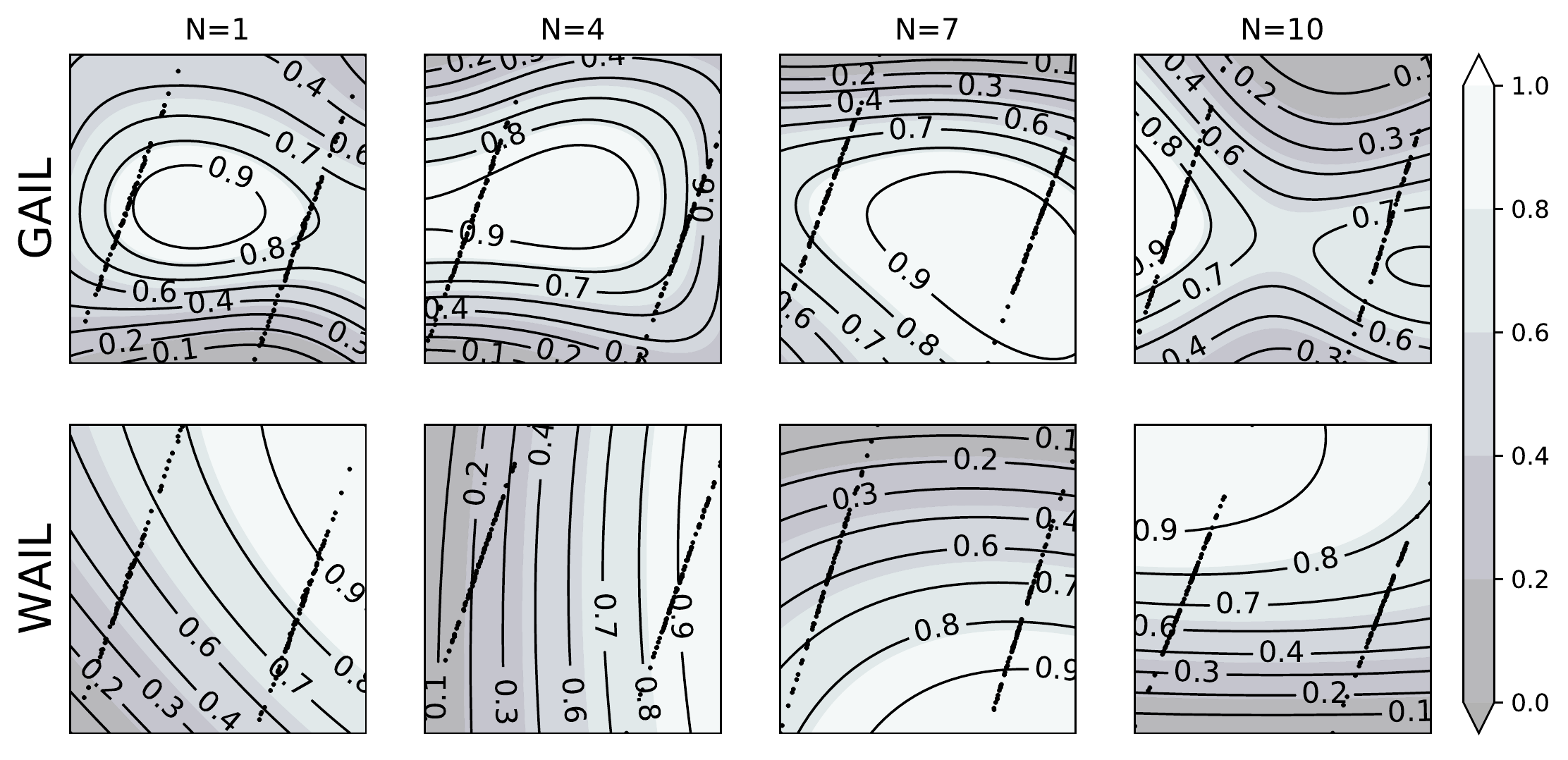}
	} \\
	\subfloat[MountainCar\label{fig:reward_surface_mountaincar}]{%
		\includegraphics[width=0.8\textwidth]{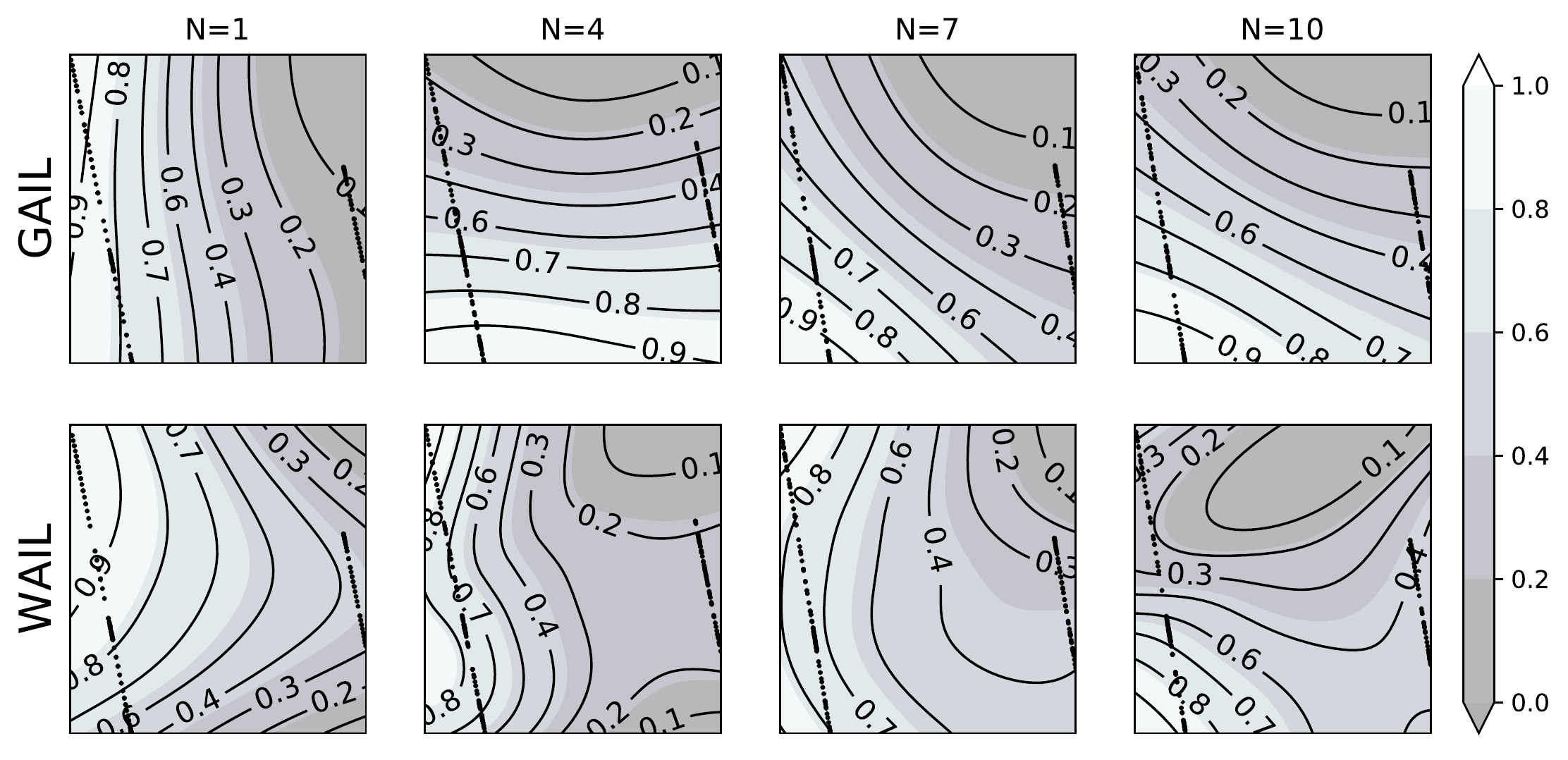}
	}\\
	\subfloat[Acrobot\label{fig:reward_surface_acrobot}]{%
		\includegraphics[width=0.8\textwidth]{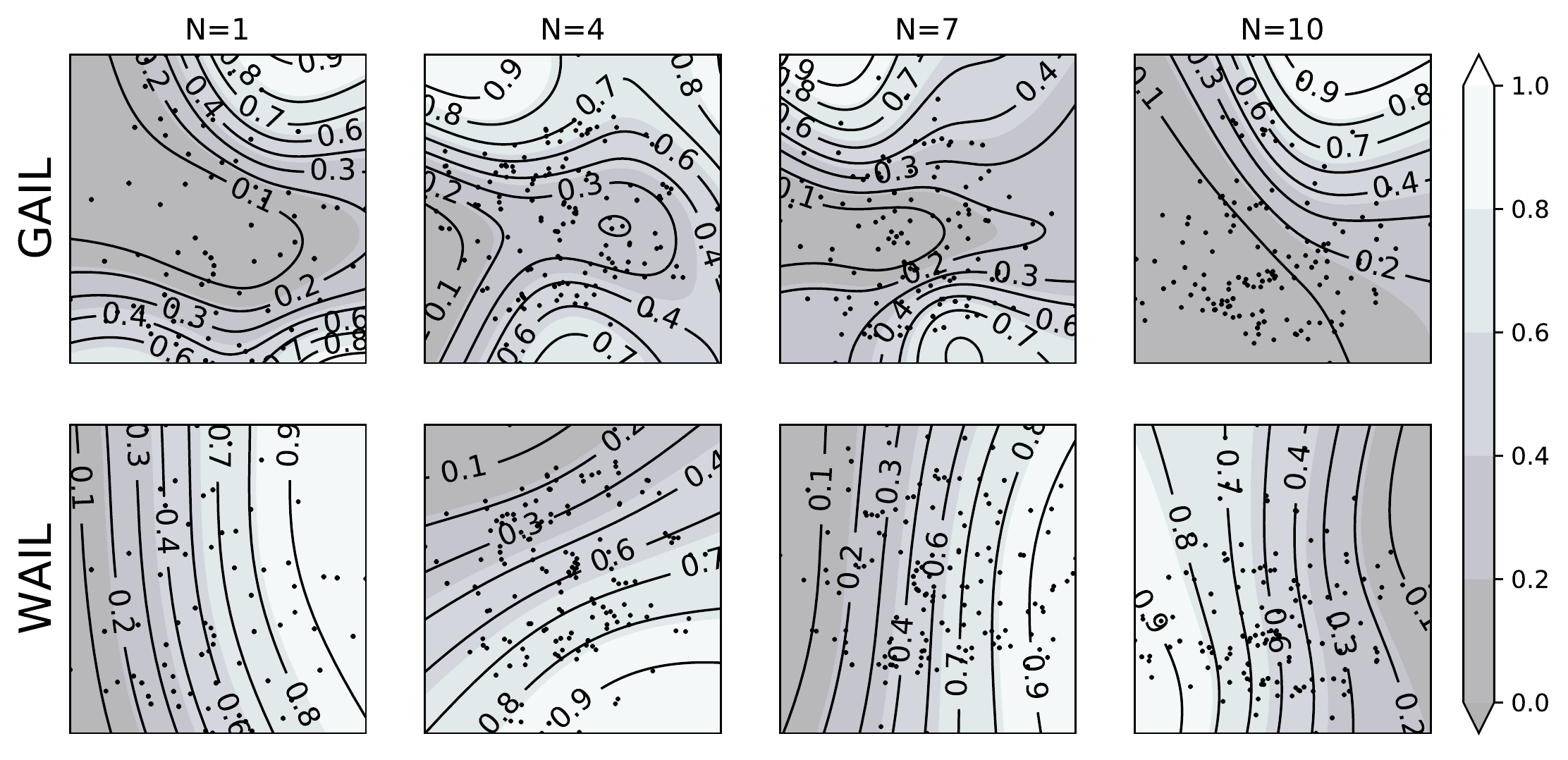}
	}
	\label{fig:reward_surface}
	\phantomcaption
\end{figure}

\begin{figure}[!ht]\ContinuedFloat
	\centering
	\subfloat[Hopper\label{fig:reward_surface_hopper}]{%
		\includegraphics[width=0.9\textwidth]{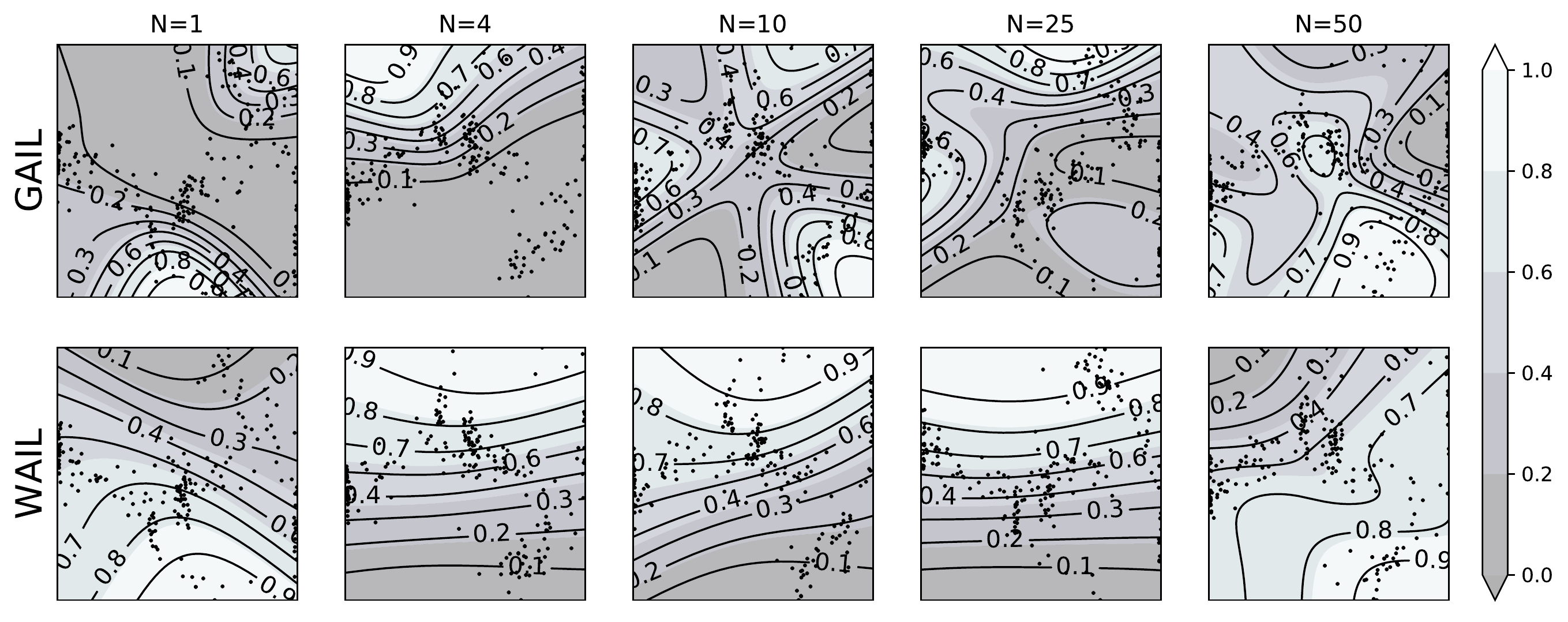}
	}\\
	\subfloat[Walker\label{fig:reward_surface_walker}]{%
		\includegraphics[width=0.9\textwidth]{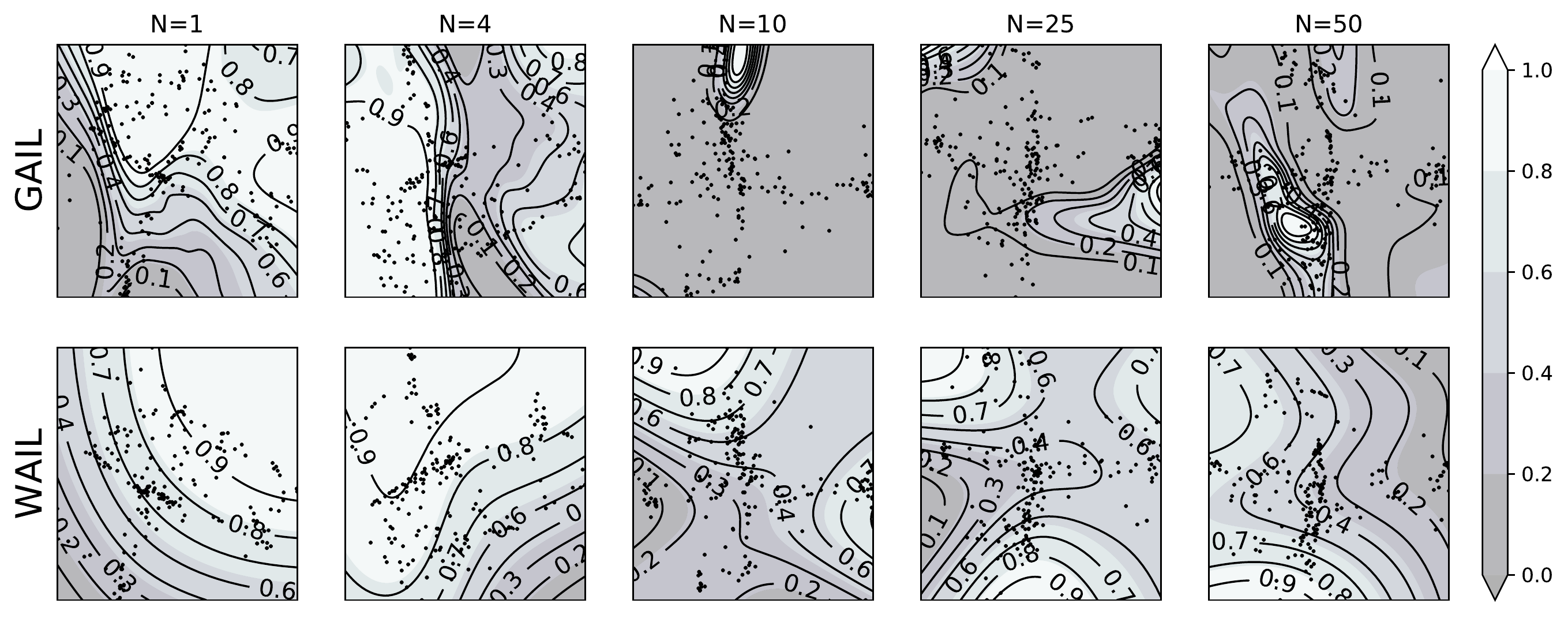}
	}\\
	\subfloat[HalfCheetah\label{fig:reward_surface_halfcheetah}]{%
		\includegraphics[width=0.9\textwidth]{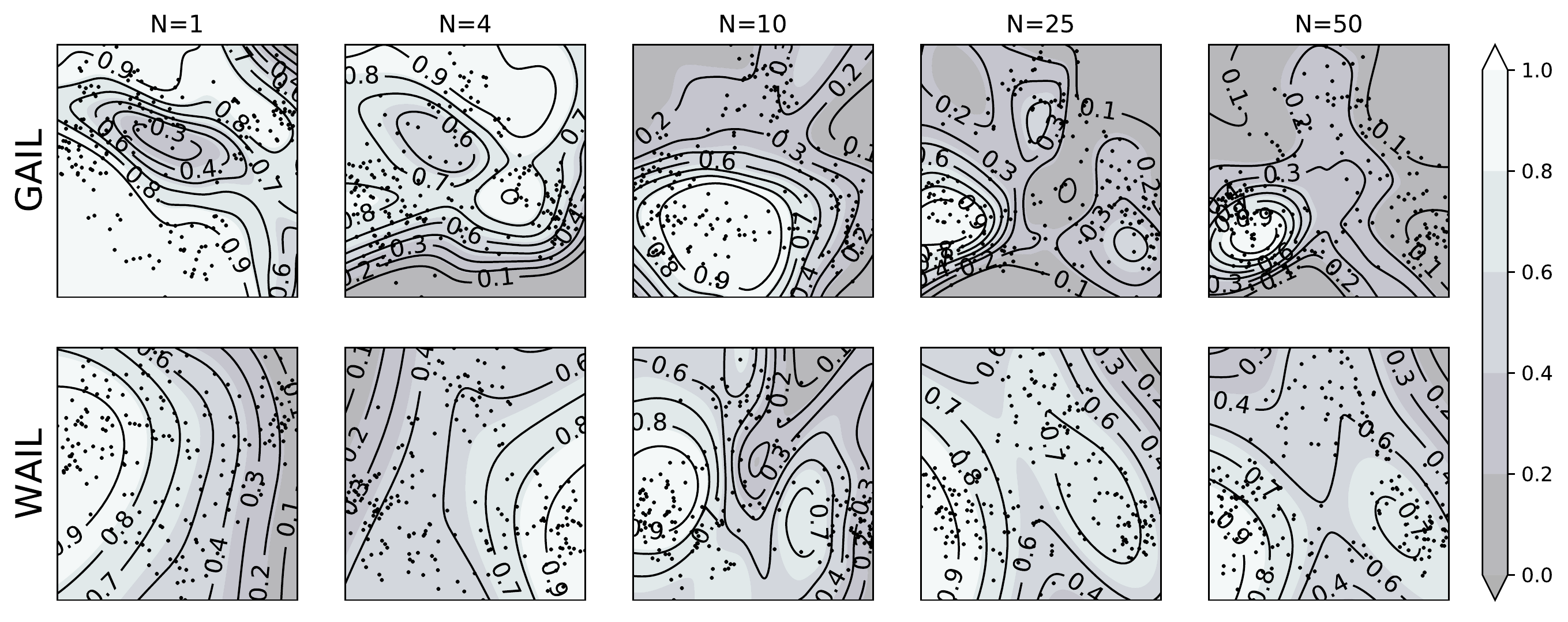}
	}
	\label{fig:reward_surface2}
	\phantomcaption
\end{figure}

\begin{figure}[!ht]\ContinuedFloat
	\centering
	\subfloat[Ant\label{fig:reward_surface_ant}]{%
		\includegraphics[width=0.9\textwidth]{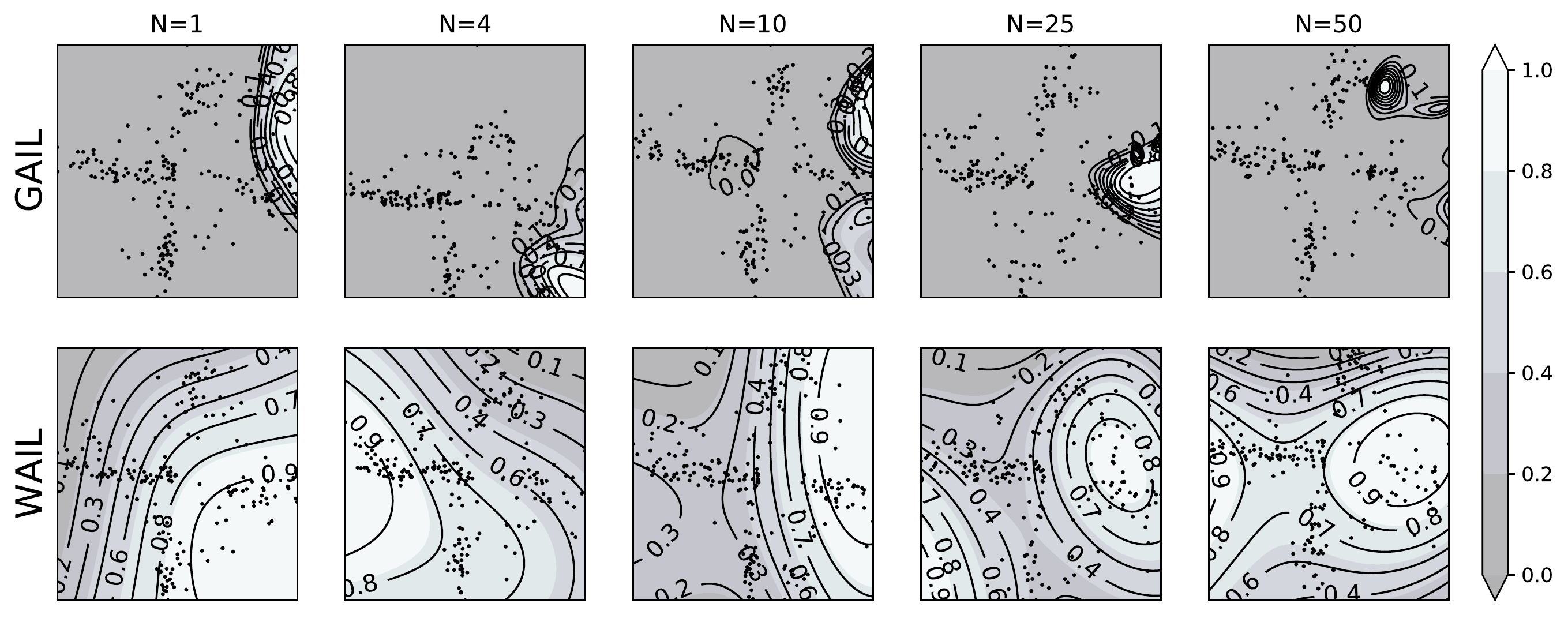}
	} \\
	\subfloat[Humanoid\label{fig:reward_surface_humanoid2}]{%
		\includegraphics[width=\textwidth]{exp_reward_humanoid.pdf}
	}\\
	\subfloat[Reacher\label{fig:reward_surface_reacher}]{%
		\includegraphics[width=\textwidth]{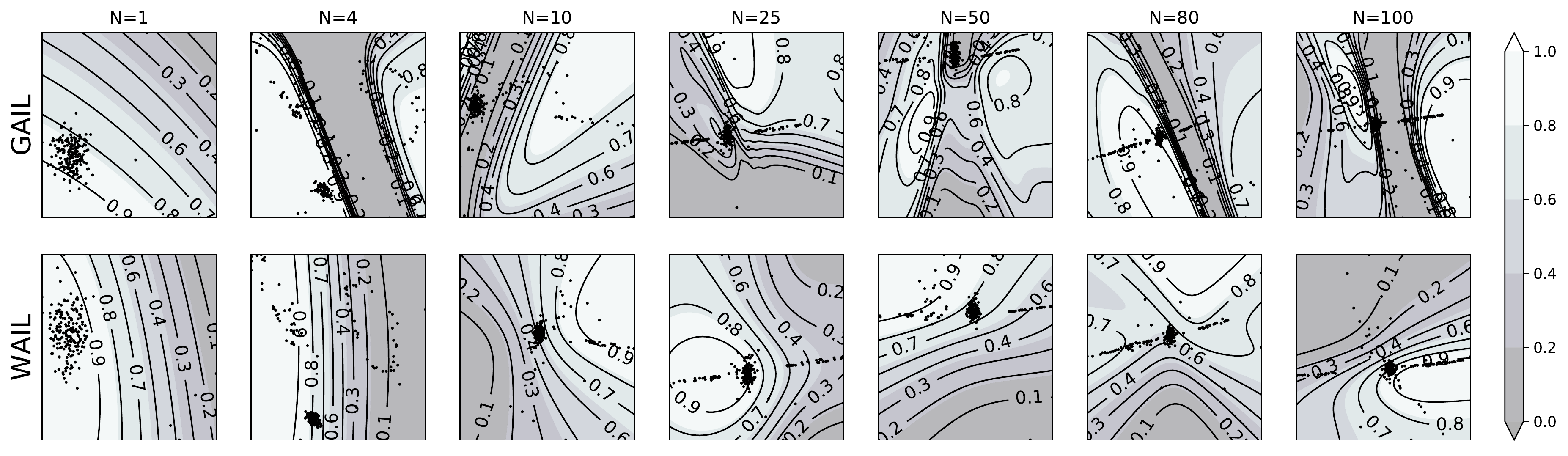}
	}
	\label{fig:reward_surface3}
	\caption{Reward surfaces of \textsc{Wail} and \textsc{Gail} on $9$ control tasks with respect to different expert data sizes.}
	
\end{figure}

\clearpage

\end{document}